%% file: main.tex
\documentclass[10pt,twocolumn,letterpaper]{article}

\usepackage[pagenumbers]{cvpr} 

\input{preamble}
\definecolor{cvprblue}{rgb}{0.21,0.49,0.74}
\usepackage[pagebackref,breaklinks,colorlinks,allcolors=cvprblue]{hyperref}
\usepackage{graphicx}
\usepackage{caption}
\usepackage{amsthm}
\usepackage{amssymb}
\usepackage{algorithm}

\usepackage{booktabs}
\usepackage{amsmath}
\usepackage{multirow} 

\usepackage[table]{xcolor}
\usepackage{makecell}
\usepackage[table]{xcolor}
\usepackage[T1]{fontenc}
\usepackage{graphicx}
\usepackage{float}
\usepackage{pifont}       
\usepackage{algpseudocode} 

\def\paperID{2041} 
\def\confName{CVPR}
\def\confYear{2026}

\newcommand{\name}{MultiCrafter}
\title{\name: High-Fidelity Multi-Subject Generation via Disentangled Attention and Identity-Aware Preference Alignment}
\definecolor{badcolor}{HTML}{FFEBEE} 
\definecolor{goodcolor}{HTML}{E8F5E9} 
\definecolor{bettercolor}{HTML}{E3F2FD} 
\author {
    Tao Wu$^{1}$\thanks{{} Equal contribution.},\hspace{.4em}
    Yibo Jiang$^{2}$\footnotemark[1],\hspace{.4em}
    Yehao Lu$^{1}$,\hspace{.3em}
    Zhizhong Wang$^{3}$,\hspace{.3em}
    Zeyi Huang$^{3}$,\hspace{.3em}
    Zequn Qin$^{2}$,\hspace{.3em}
    Xi Li$^{1}$\thanks{{} Corresponding authors.}
    \\
    $^1$College of Computer Science and Technology, Zhejiang University \\
    $^2$School of Software Technology, Zhejiang University \quad
    $^3$Huawei Technologies Ltd
}
\begin{document}
\Crefname{figure}{Fig.}{Figs.}

\Crefname{table}{Tab.}{Tabs.}

\Crefname{section}{Sec.}{Secs.}
\Crefname{equation}{Eq.}{Eqs.}
\twocolumn[{
\renewcommand\twocolumn[1][]{#1}
\maketitle
\begin{center}
    \centering
    \vspace*{-.8cm}
    \includegraphics[width=0.9\textwidth]{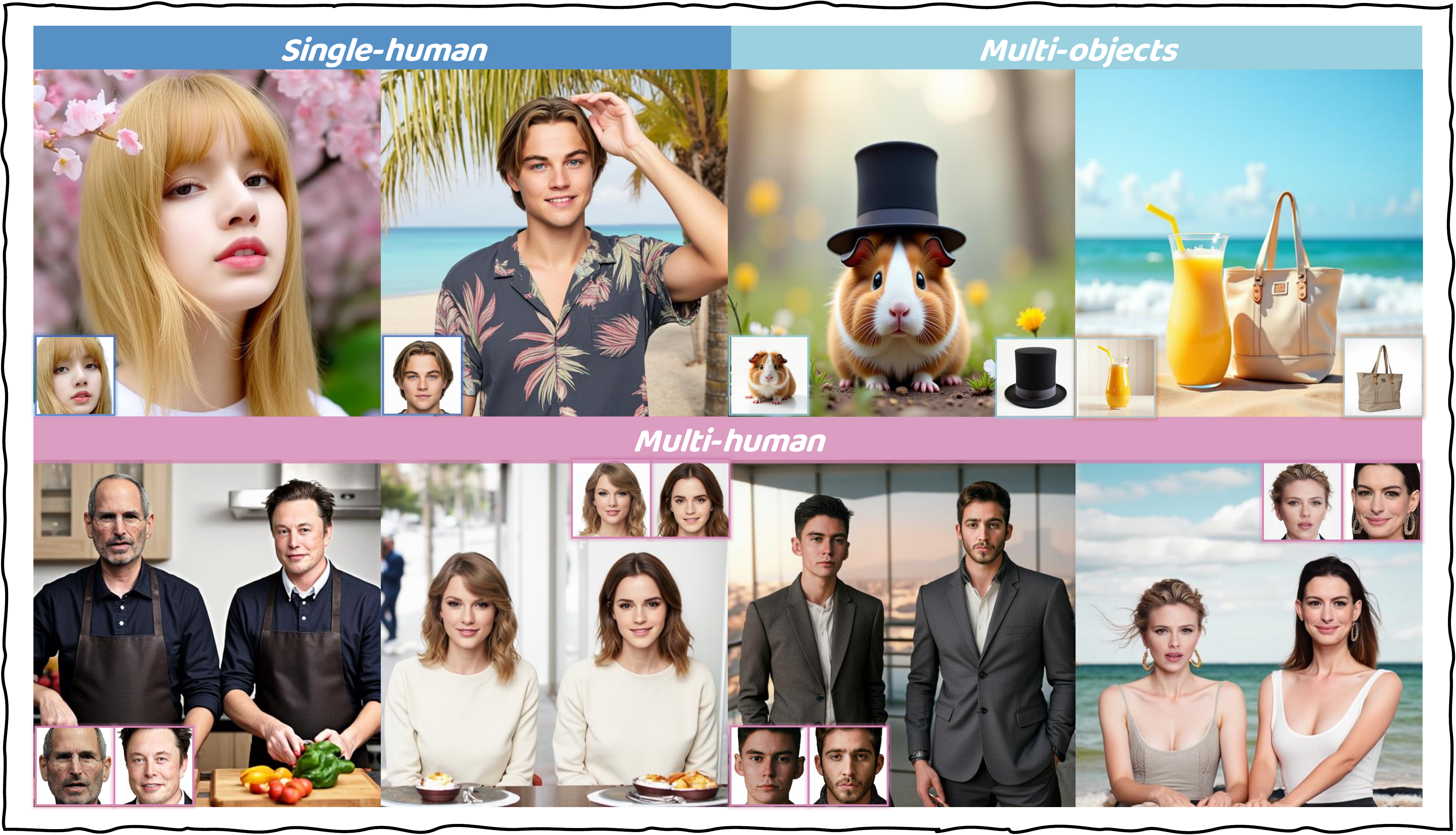}
    \captionof{figure}{MultiCrafter enables high-quality multi-subject personalization. Inputs are surrounded by squares.}
\label{fig:teaser}
\end{center}
}]
\footnotetext[1]{$^*$ These authors contributed equally. $^\dagger$ Corresponding author.}
\input{sections/0_abstract.tex}
\input{sections/1_introduction.tex}
\input{sections/2_related_work.tex}
\input{sections/3_preliminary.tex}

\input{sections/4_methods.tex}

\input{sections/5_experiments.tex}
\input{sections/6_conclusion.tex}

{
    \small
    \bibliographystyle{ieeenat_fullname}
    \bibliography{main}
}

\input{sections/X_appendix}

\end{document}

%% file: sections/0_abstract.tex
\begin{abstract}
Multi-subject image generation aims to synthesize user-provided subjects in a single image while preserving subject fidelity, ensuring prompt consistency, and aligning with human aesthetic preferences.
Existing In-Context-Learning based methods are limited by their highly coupled training paradigm. 
These methods attempt to achieve both high subject fidelity and multi-dimensional human preference alignment within a single training stage, relying on a single, indirect reconstruction loss, which is difficult to simultaneously satisfy both these goals.
To address this, we propose MultiCrafter, a framework that decouples this task into two distinct training stages. 
First, in a pre-training stage, we introduce an explicit positional supervision mechanism that effectively resolves attention bleeding and drastically enhances subject fidelity.
Second, in a post-training stage, we propose Identity-Preserving Preference Optimization, a novel online reinforcement learning framework. 
We feature a scoring mechanism to accurately assess multi-subject fidelity based on the Hungarian matching algorithm, which allows the model to optimize for aesthetics and prompt alignment while ensuring subject fidelity achieved in the first stage.
Experiments validate that our decoupling framework significantly improves subject fidelity while aligning with human preferences better.
\end{abstract}

%% file: sections/1_introduction.tex
\section{Introduction}
\label{sec:intro}
\vspace{-0.2cm}
With improvements in data quality and the large-scale adoption of Diffusion Transformers (DiT)~\cite{flux2024,esser2024scaling,gong2025seedream2,gao2025seedream3,li2024hunyuandit}, text-to-image models have developed rapidly, driving demand for personalized generation.
Multi-subject image generation, which aims to create images featuring multiple user-provided subjects, is a challenging but significant task for personalized generation.
This task must simultaneously satisfy two distinct and challenging objectives:
(\textit{i}) achieving high subject fidelity for multiple subjects, and 
(\textit{ii}) aligning with human preferences, encompassing aesthetic quality, semantic accuracy, and precise text alignment.

Mainstream In-Context-Learning (ICL)~\cite{huang2024group,huang2024context,tan2024ominicontrol} based works, such as UNO \cite{wu2025less} and OmniGen \cite{xiao2025omnigen,wu2025omnigen2}, have demonstrated a high degree of adaptability to the transformer architecture, significantly improving the quality of multi-subject generation. 
However, despite their efforts, these methods frequently suffer from suboptimal results, such as low subject fidelity caused by attribute leakage between subjects. 
This raises a critical question: \textit{why do existing ICL-based methods falter in the multi-subject setting?}

\begin{figure*}[tb]
    \centering
    \includegraphics[width=0.98\linewidth]{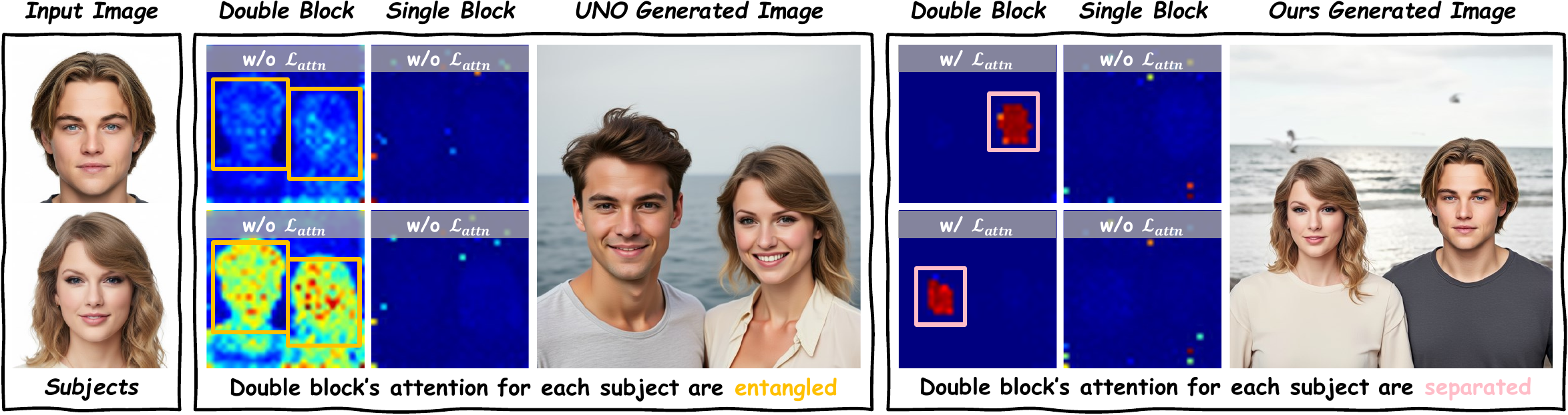}
    \vspace{-0.2cm}
    \caption{Visual comparison of attention maps. The ICL-based method UNO (left), fails to preserve subject fidelity. This is due to attention bleeding, where the double block's attention regions for each subject are entangled, leading to attribute leakage. Our method overcomes this problem and maintains subject fidelity.}
    \label{fig:intro}
    \vspace{-0.5cm}
\end{figure*}

\textbf{It is difficult to achieve these two challenging objectives simultaneously using only a highly coupled training paradigm.}
Specifically, these methods attempt to simultaneously achieve high subject fidelity and align with multi-dimensional human preferences, relying solely on a single and indirect reconstruction loss during a training stage. 
We argue that these two distinct learning objectives should not be forced into a single supervision signal, as this would inevitably force the model to find a sub-optimal result between these two objectives.
This coupled learning objective produces two major failures.
First, for subject fidelity, it fails to resolve the underlying spatial entanglement between subject features and their intended locations. 
As shown in \Cref{fig:intro}, this leads to the "attention bleeding" phenomenon (e.g., in UNO, the double block's attention regions for each subject are entangled), causing attribute leakage and significantly harming fidelity.
Second, for human preference, these methods suffer from a severe proxy-objective mismatch, as the reconstruction loss cannot directly align multi-dimensional human preferences.

In this work, we introduce MultiCrafter, a multi-subject generation framework that employs a divide-and-conquer strategy to decouple this complex process by learning subject fidelity and human preference alignment separately at different training stages.
We implement this via two stages focusing on different objectives: a fidelity-focused pre-training stage and a preference-optimizing post-training stage.
Specifically, in the pre-training stage, we focus on achieving high subject fidelity. 
To address the attribute leakage caused by attention bleeding, we propose an \textbf{Identity-Disentangled Attention Regularization}. 
This mechanism applies explicit positional supervision only during the training phase to double blocks in FLUX~\cite{flux2024}, which are pivotal regions for feature injection and spatial control. 
During training, we add this explicit supervision to the standard reconstruction loss, guiding the model's attention to concentrate on the precise spatial layout of each subject in the generated image. 
This compels the model to distinguish between different subject features and learn distinct, disentangled attention regions for each subject, drastically reducing attribute leakage. 
However, this regularization requires the model to master the complex and diverse spatial layouts caused by different prompts and subjects. 
Finding that this challenge exceeds the capacity of a single LoRA module, we use a Mixture-of-Experts LoRA (MoE-LoRA) to provide the required specialized capacity

In the post-training stage, to align with multi-dimensional human preferences, we design a novel online reinforcement learning framework. 
We introduce a stable \textbf{Identity-Preserving Preference Optimization} that directly aligns the model across aesthetic quality, text-image alignment, and subject fidelity. 
To accurately measure subject fidelity, we introduce a Multi-ID Alignment Reward, which uses the Hungarian matching algorithm to maximize the overall match quality between multiple generated subjects and their references for precise scoring. 

Experiments demonstrate that our approach, which decouples the learning of subject fidelity and human preference optimization into distinct training stages, achieves significant improvements over existing methods. 
Our main contributions are as follows:

\begin{itemize}
    \item We propose explicit positional supervision that disentangles attention across subjects, thereby reducing attribute leakage and enhancing subject fidelity.
    
    \item To our knowledge, we design the first online reinforcement learning framework tailored for multi-subject generation, introducing a Multi-ID Alignment Reward for stable and preference-aligned training.

    \item We demonstrate improvements over existing methods, particularly in subject fidelity, achieving state-of-the-art performance in overall evaluations.
\end{itemize}

%% file: sections/2_related_work.tex
\section{Related Work}
\label{sec:related_work}


\noindent\textbf{Subject-driven Generation} has attracted increasing attention~\cite{chen2023disenbooth, han2023svdiff, shi2023instantbooth, ruiz2024hyperdreambooth, hua2023dreamtuner, han2024face, liu2023cones, liu2023cones2,feng2025personalize, wang2024high, chen2024disenstudio}. 
These works can be broadly categorized into two types.
(1) Fine-tuning-based methods~\cite{gal2022image,ruiz2023dreambooth,kumari2023multi}. These approaches achieve customization by fine-tuning part of the model's parameters.
(2) Tuning-free methods~\cite{ye2023ip,wang2024instantid,li2024photomaker}. These approaches leverage large-scale training to eliminate the need for retraining when the subject changes. 
Recently, the powerful capabilities of foundation models based on the DiT \cite{peebles2023scalable} architecture have greatly enhanced the generation of multiple subjects, resulting in many excellent multi-subject generation works, \eg, HunyuanCustom~\cite{hu2025hunyuancustom}, OmniControl \cite{tan2024ominicontrol}, UniReal \cite{chen2025unireal}, UNO \cite{wu2025less}, OmniGen~\cite{xiao2025omnigen}.
Some methods~\cite{liu2023customizable,gu2024mix,wang2025msdiffusion} address attribute leakage using user-provided spatial layouts to constrain attention during inference, which increases complexity for users. 
Our ICL-based method directly constrains attention between the reference and generated images during training, bypassing the text modality, enabling precise spatial alignment and requiring no extra user input at inference.

\noindent\textbf{Reinforcement Learning for Text-to-Image Generation} has become an active area of research. Initial strategies included policy gradient methods like Proximal Policy Optimization (PPO)~\cite{schulman2017proximal, black2023training, fan2024reinforcement, gupta2025simple, miao2024training, zhao2025score, xiao2025fastcomposer}. A subsequent major development is the adoption of Direct Preference Optimization (DPO) and its variants~\cite{wallace2024diffusion, yang2024using, yuan2024self, liu2025videodpo, zhang2024onlinevpo, furuta2024improving,li2025magicid}. 
Some recent works introduce online RL technology, Group Relative Policy Optimization (GRPO) ~\cite{shao2024deepseekmath}, into Text-to-Image Generation, achieving significant performance gains. Flow-GRPO~\cite{liu2025flow}, DanceGRPO~\cite{xue2025dancegrpo} introduce exploration by reformulating the deterministic Ordinary Differential Equation (ODE) of flow-matching models into a Stochastic Differential Equation (SDE). 
The improved MixGRPO~\cite{li2025mixgrpo} further boosts training efficiency with a mixed ODE-SDE framework. 
However, these methods have been limited to the basic text-to-image task, leaving the application of online reinforcement learning to multi-subject driven generation largely unexplored.

%% file: sections/3_preliminary.tex
\section{Preliminary}
\label{sec:preliminary}
\noindent\textbf{Flow Matching}. 
Flow Matching~\cite{lipman2022flow} is gradually replacing DDPM as the mainstream for text-to-image models due to its efficient sampling strategies.
These models typically first train an autoencoder (an encoder ${\cal E}$ and a decoder ${\cal D}$) to obtain the latent space representation $z_0 = {\cal E}(x)$ of an image $x$.
Let $z_0$ be a data sample, $\epsilon\in \mathcal{N}(0,1)$ is the Gaussian noise, and $c_{text}$ be the text prompt. 
Flow Matching formulates generation as a continuous transformation along an Ordinary Differential Equation (ODE), $\frac{dz_t}{dt} = v(z_t, t),\ t \in [0,1]$, which deterministically maps noise to data. The interpolated data at time $t$ is
$z_t = (1 - t)z_0 + t \epsilon.$
A neural network $v_\theta(x_t, t)$ is trained to approximate the velocity field of this ODE, with the objective
\begin{equation} \label{eq:2}
\mathcal{L}_{diff} = \mathbb{E}_{t, z_0, \epsilon\in \mathcal{N}(0,1)} \|v - v_\theta(z_t, t, c_{text})\|^2. 
\end{equation}
The DiT framework and Flow Matching are widely used in recent diffusion models, such as Stable Diffusion3~\cite{esser2024scaling} and Flux~\cite{flux2024}.
In this paper, we use Flux as our base model.

\noindent\textbf{Group Relative Policy Optimization (GRPO)}. GRPO struggles with ODE-based flow models because their deterministic systems lack the inherent stochasticity essential for reinforcement learning. So DanceGRPO~\cite{xue2025dancegrpo} and Flow-GRPO~\cite{liu2025flow} convert ODEs to SDEs, enabling stochastic reinforcement learning for image generation. MixGRPO~\cite{li2025mixgrpo} further improves efficiency by applying this only within a sliding time window during training. Given the prompt $c_{text}$, the training process in MixGRPO is similar to Flow-GRPO and DanceGRPO, but only optimizes the time steps sampled within the interval $S$. The behavior of this window is governed by key hyperparameters: the window size $w$ sets the number of consecutive timesteps to optimize at once, the shift interval $\tau$ determines how many training iterations pass before the window moves, and the window stride $s$ specifies how many timesteps the window advances during a shift. The final training objective is given by:
\begin{small}
\begin{equation} \label{eq:8}
\begin{aligned}
J(\theta) = \mathbb{E}_{x \sim \pi_{\theta_\text{old}}(\cdot|c)} & \Bigg[ \frac{1}{N} \sum_{i=1}^N \frac{1}{|S|} \sum_{t \in S}  \min \Big( r_i^t(\theta) A_i, \\ &clip(r_i^t(\theta), 1-\beta, 1+\beta) A_i \Big) \Bigg],
\end{aligned} 
\end{equation}
\end{small}
\hspace{-2.5mm} where $ r_i^t(\theta) = \frac{\pi_\theta(x_{t+1} \mid x_t, c)}{\pi_{\theta_\text{old}}(x_{t+1} \mid x_t, c)}$ is the policy ratio and $A_i$ is the advantage score. $\beta$ is a hyperparameter that serves to clip the policy ratio, ensuring stable updates, and
\begin{equation} \label{eq:9}
\quad A_i = \frac{R(x_i^T, c) - mean\{R(x_i^T, c)\}_{i=1}^N}{std\{R(x_i^T, c)\}_{i=1}^N},
\end{equation}
where $R(x_i^0, c)$ is provided by the reward model.

%% file: sections/4_methods.tex
\section{Methods}
\label{sec:methods}

\begin{figure*}[tb]
    \centering
    \captionsetup{font={small}} 
    \includegraphics[width=0.96\linewidth]{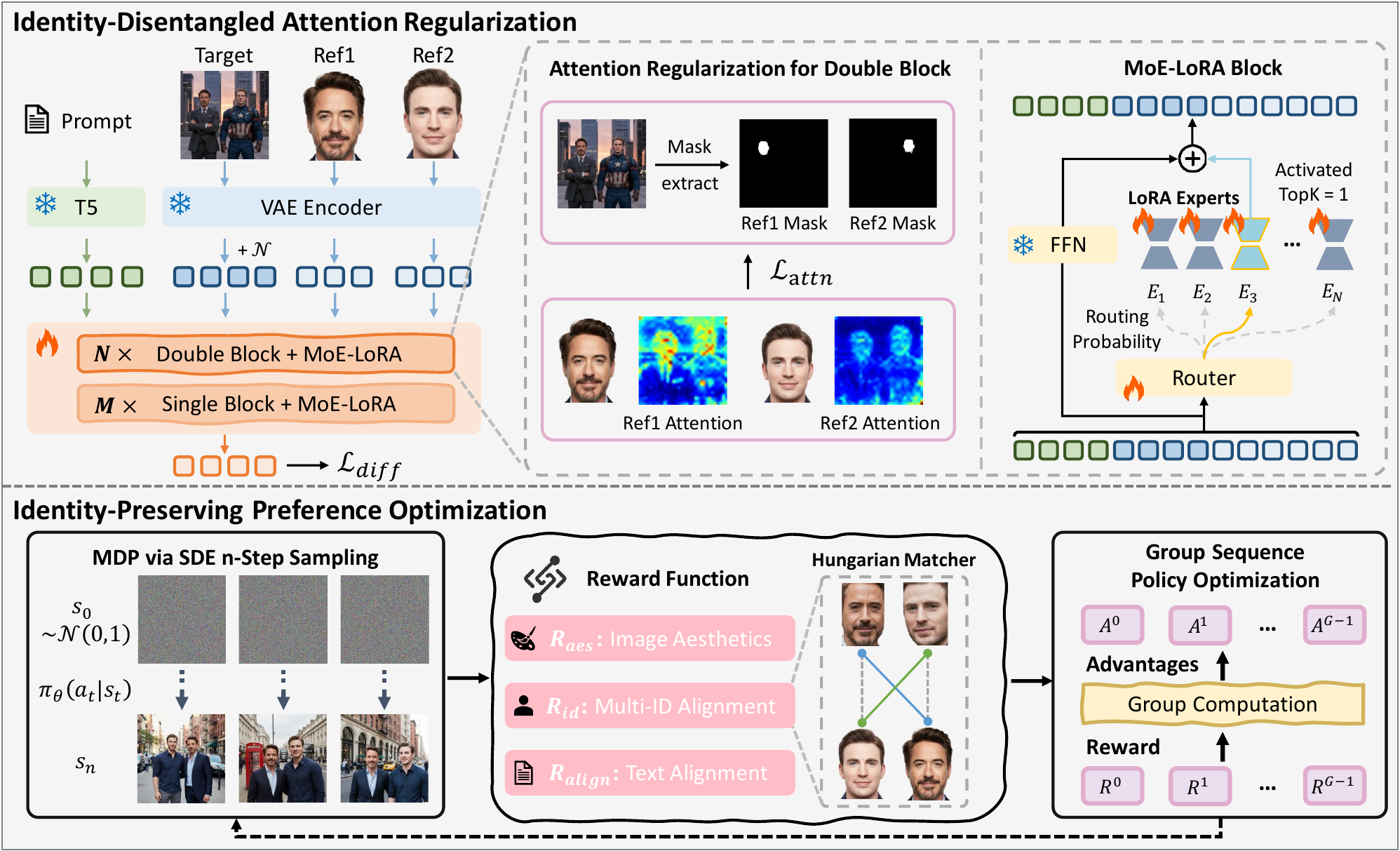}
    \caption{Overall pipeline of MultiCrafter. Our framework is built on two core innovations: (Top) Identity-Disentangled Attention Regularization uses positional supervision to prevent attribute leakage and the MoE-LORA architecture to boost model capacity for diverse scenarios; and (Bottom) the Identity-Preserving Preference Alignment framework employs a novel online reinforcement learning strategy with a Multi-ID Alignment Reward and the stable GSPO algorithm to align the model with human preferences.}
    \label{fig:pipline}
    \vspace{-0.5cm}
\end{figure*}


In this section, we first explore the impact of highly coupled training objectives on the entire training process and present the overall architecture of our decoupled framework in \Cref{sec:over}.
Then, in \Cref{sec:loss}, we introduce \textbf{Identity-Disentangled Attention Regularization}, which we used during pre-training to achieve the goal of high subject fidelity.
Finally, \Cref{sec:rl} describes \textbf{Identity-Preserving Preference Optimization (IPPO)}, which we used in the post-training stage to directly align the model with multi-dimensional human preferences.

\subsection{Explore the In-Context-Learning}
\label{sec:over}

Given $N$ reference images of different subjects, we aim to generate images of these subjects according to a text prompt. Existing ICL-based methods encode these subjects into latent features, $\mathcal{Z} = \{\mathbf{z}_{i}^{ref}\}_{i=1}^{N}$, and leverage DiT-based attention mechanisms for customized generation.
During training, these methods attempt to achieve high subject fidelity and conform to human preferences simultaneously, using a single reconstruction loss during the training phase.
This forces the model to find a sub-optimal compromise between these two objectives, which especially reduces subject fidelity when subjects have similar attributes.
Furthermore, this simple supervision method fails to adequately account for multi-dimensional human preferences.
Therefore, we decouple the training process into pre-training and post-training stages, allowing the model to focus separately on subject fidelity and alignment with human preferences.

To explore how to improve subject fidelity during pre-training, we visualized the attention maps of a representative method~\cite{wu2025less}, focusing on the double and single block structures. As shown on the left of~\Cref{fig:intro}, we identify two key phenomena. First, the double blocks of Flux are far more pivotal in determining the spatial layout of the reference subject than the single block. 
Second, methods trained solely on a reconstruction loss lead to an undesired entanglement between subject-specific attention fields, causing attribute leakage and severely compromising subject fidelity. 
We can therefore infer a critical condition for high fidelity: the peak response within the double block's attention scores, corresponding to a specific subject, must consistently align with that subject's spatial region in the generated output.
We leverage this in our method.

\subsection{\fontsize{10.8}{13}\bfseries\selectfont Identity-Disentangled Attention Regularization}
\label{sec:loss}
Based on ~\Cref{sec:over}, the subject fidelity can be enhanced by strictly aligning the hot spot areas of the attention scores of double blocks with the spatial positions of the corresponding subjects in the generated image. We design a simple but effective regularization. 
Specifically, during the training process, for the latent space feature $\mathbf{z}_{i}^{ref}$ of the $i$-$th$ reference subject, we partition the reference feature $\mathbf{z}_{i}^{ref}$ into patches and apply positional encoding, resulting in a sequence of 1D tokens $\mathbf{z}_{i}^{r\prime}\in\mathbb{R}^{l \times c}$, $l$ is the number of tokens, $c$ is the number of channels.
Then, we can obtain its attention map with the generated content $z_t$ at the $k$-th double block:
\begin{equation}
    m_{k}^{i}= \operatorname{Softmax}(\frac{\mathbf{Q}_{k,i}\mathbf{K}_{k}^{T}}{\sqrt{d}}),
\end{equation}
where $\mathbf{Q}_{k,i}\in \mathbb{R}^{l \times c}$ is the query generated from the $i$-th subject image within the $k$-th double block, $\mathbf{K}\in \mathbb{R}^{l_t\times c}$ is the key produced by the noisy image latent tokens in the current layer, and $l_t$ denotes the number of tokens in the noisy image latent.
For a model with $N$ double blocks, we can obtain the attention maps corresponding to the $i$-th subject from all blocks, which are aggregated into a set $\{m_1^{i}, m_2^{i}, ..., m_N^{i}\}$. We then average and normalize this set to obtain the mean attention map $\hat{M}_{i}$.
By pre-annotating the training data, we obtain the ground-truth mask ${M}_{i}$ corresponding to the $i$-th subject within the generated image. 
Notably, for human subjects, we exclusively use the facial region as the reference image and similarly focus only on the facial area for the generated image.

We employ the dice loss~\cite{milletari2016v}, a standard loss function for segmentation tasks, to minimize the discrepancy between each ground-truth mask ${M}_{i}$ and the corresponding mean attention map $\hat{M}_{i}$.
The formulation is as follows:
\begin{equation}
\label{eq:dice_loss}
\mathcal{L}_{attn} = \sum_{i=1}^{N} \left( 1 - \frac{2 \sum_{j} (\hat{M}_{i,j} \cdot M_{i,j}) + \epsilon}{\sum_{j} \hat{M}_{i,j} + \sum_{j} M_{i,j} + \epsilon} \right),
\end{equation}
where the index $j$ iterates over all spatial locations of the maps, and $\epsilon$ is a small constant added for numerical stability.
By minimizing this attention regularization loss $\mathcal{L}_{attn}$, we explicitly encourage the model's attention mechanism to concentrate on the precise spatial regions occupied by each subject. This forces a spatial disentanglement of subjects within the attention maps.
This avoids attribute leakage and improves the subject fidelity of the generated image. 
The final loss function of our framework is defined as ($\lambda$ is a factor that balances the loss weight):
\begin{equation}
    L = L_{diff}+\lambda\cdot L_{attn}.
\end{equation}

However, for efficient training, existing ICL-based methods usually use LoRA to fine-tune the model.
But the significant variance in the spatial layouts generated by different prompts and subjects poses a challenge for a standard LoRA, whose limited capacity is insufficient to master the complex task. 
To address this, and inspired by the success of MoE-LoRA in multitask tuning~\cite{feng2024mixture,liu2024moe,gou2023mixture}, we adopt an MoE-LoRA architecture to expand model capacity. 
This allows different experts to focus on the spatial layout of various scenarios, effectively alleviating this problem.
We strategically integrate the MoE-LoRA into the output feed-forward network (FFN) layers of the Flux, while other layers are adapted using standard LoRA for parameter efficiency.
Since the use of experts in this situation is difficult to define manually, we choose to let the model learn this implicitly.
Specifically, given an input vector $h$ to the FFN layer, we define the number of experts as $N_e$, the rank of each LoRA as $r$, and the scaling factor as $\alpha$.
A lightweight gating network, $g_\theta$, which dynamically routes the input vector $h$ to the most suitable experts, computes a vector of logits, $p \in \mathbb{R}^{N_e}$, for the experts:
\begin{equation}
p = \operatorname{Softmax}(\operatorname{TopK}(W_g \cdot h, k)),
\end{equation}
where $W_g \in \mathbb{R}^{N_e \times d_{in}}$ is the weight of the gating network. $\operatorname{TopK}(\cdot, k)$ enforces sparsity by retaining only the top $k$ logit values and masking the others to $-\infty$, thus activating only a small subset of experts.
Each of the $N_e$ experts is an independent LoRA module, parameterized by matrices $W_A^i \in \mathbb{R}^{r \times d_{in}}$ and $W_B^i \in \mathbb{R}^{d_{out} \times r}$. The final output of the MoE-LoRA layer, $h_{out}$, is computed by adding the weighted sum of the selected experts' outputs to the output of the original FFN layer:
\begin{equation}
h_{out} = \text{FFN}(h) + \sum_{i=1}^{N_e} p_i \cdot \left(\frac{\alpha}{r} \cdot W_B^i \cdot W_A^i \cdot h\right).
\end{equation}

\subsection{Identity-Preserving Preference Optimization}
\label{sec:rl}
To further enhance the generation quality and align with human preferences, we introduce a post-training stage using reinforcement learning. 
This aims to refine aesthetic appeal and text-image alignment without compromising the subject fidelity.
We adapt the efficient MixGRPO framework~\cite{li2025mixgrpo}, which confines stochastic optimization to a sliding window $S$. 
However, standard GRPO with its token-level policy ratios can exhibit instability, particularly when training MoE models due to expert routing fluctuations~\cite{zheng2025group}.
To mitigate this and better suit our MoE-LoRA architecture from \Cref{sec:loss}, we replace the GRPO objective with the more stable Group Sequence Policy Optimization (GSPO)~\cite{zheng2025group}.
Specifically, we replace the token-level policy ratio in \Cref{eq:9} with a sequence-level policy ratio $s_i(\theta)$ defined over the denoising steps within the sliding window $S$:
\begin{equation}
\label{eq:gspo_ratio}
s_i(\theta) = \exp\left(\frac{1}{|S|}\sum_{t\in S} \log \frac{\pi_{\theta}(x_{t+1}|x_t, c, \mathcal{Z})}{\pi_{\theta_{old}}(x_{t+1}|x_t, c, \mathcal{Z})}\right).
\end{equation}
This sequence-level ratio reflects the overall policy shift for the entire sequence within the optimization window, leading to more stable gradients.
The advantage score $A_i$ is calculated as in \Cref{eq:9}, but using our composite reward from \Cref{eq:reward_total}.
The final optimization objective is thus:
\begin{small}
\begin{equation}
\label{eq:mixgspo_obj}
\begin{aligned}
J(\theta) = \mathbb{E}_{x \sim \pi_{\theta_\text{old}}(\cdot|c, \mathcal{Z})} &\left[ \frac{1}{N} \sum_{i=1}^N \min \Big( s_i(\theta) A_i, \right. \\ 
&\left. clip(s_i(\theta), 1-\beta, 1+\beta) A_i \Big) \right].
\end{aligned}
\end{equation}
\end{small}


Besides, a good reward model is very important for online reinforcement learning. We construct a reward model based on three dimensions: aesthetics, text alignment, and subject fidelity.
The total reward $R(x_i^T, c, \mathcal{Z})$ for a generated image $x_i^T$ given a text prompt $c_{text}$ and a set of reference subject latents $\mathcal{Z}$ is a weighted sum of three scores:
\begin{equation}
\label{eq:reward_total}
R(x_i^T, c, \mathcal{Z}) = w_{text} R_{text} + w_{aes} R_{aes} + w_{id} R_{id},
\end{equation}
where $R_{text}$ is the text alignment reward from a pre-trained CLIP model, $R_{aes}$ is an aesthetic reward from a predictor like HPSv2~\cite{wu2023human}, $R_{id}$ is used to evaluate subject fidelity, and $w_{id}, w_{text}, w_{aes}$ are their corresponding weights, .
To accurately measure the subject fidelity of multi-subject generation results, we built the Multi-ID Alignment Reward using the Hungarian matching algorithm.
For humans, we first employ a face detector~\cite{deng2020retinaface} to extract facial embeddings from each reference image.
We then apply the same detector to the generated image to identify all faces and extract their embeddings. 
Then we construct a pairwise similarity matrix $C$ where $C_{ij}$ is the cosine similarity between the embedding of the $i$-th reference face and the $j$-th detected face. The Hungarian algorithm is then used to solve the assignment problem by finding an assignment matrix $X \in \{0, 1\}^{N_{ref} \times N_{gen}}$ that maximizes the total similarity:
\begin{equation} 
\label{eq:hungarian}
\max_{X} \sum_{i=1}^{N_{ref}} \sum_{j=1}^{N_{gen}} C_{ij} X_{ij} \quad \text{s.t.} \sum_{j=1}^{N_{gen}} X_{ij} \le 1, \sum_{i=1}^{N_{ref}} X_{ij} \le 1.
\end{equation}
where $N_{ref}$ and $N_{gen}$ are the number of reference and generated faces, respectively. 
These ensure each face is matched at most once, preventing reward hacking, stopping the model from using attribute leakage to generate multiple ``average faces" for an unearned high reward. 
For object subjects, each reference object is pre-annotated with a text prompt. We leverage Florence-2~\cite{xiao2024florence} and SAM2~\cite{ravi2024sam} to locate the corresponding object in the generated image. 
Then we compute the cosine similarity between the DINOv2~\cite{oquab2023dinov2} embeddings of the segmented region and the reference object.
We provide the complete pseudocode in the Appendix.

%% file: sections/5_experiments.tex
\section{Experiments}
\label{sec:experiments}

\subsection{Experimental Setup}
\input{tables/sota}
\noindent\textbf{Implementation Details.} 
\label{sec:imp_details}
To achieve accurate multi-subject customized generation, our training process is decoupled into two stages. Following \citep{wu2025less}, we set the resolution of generated images to $512 \times 512$ , the resolution of reference images to $320 \times 320$, and each LoRA module's rank to $r=512$. For the MoE-LoRA applied to the FFN layers, we configure it with 4 experts and activate 1 expert per forward pass. 
For reinforcement learning,  we configure with a sampling step 16, a window size of $w=2$, a shift interval of $\tau=50$, and a window stride of $s=1$. More detailed settings can be found in the Appendix.

\noindent\textbf{Datasets and BenchMark.} 
\label{sec:datasets}
We constructed separate datasets for multi-human and multi-object generation. 
Due to the scarcity of public multi-human customization datasets with adequate annotations, we designed a data collection pipeline to curate a new dataset from OpenHuman-Vid datasets~\cite{li2025openhumanvid}. We supplemented the details in the Appendix.
A total of 200k pairs of cross-pair data are obtained for multi-human generation training.
For the reinforcement learning, we curated a face collection of 80 celebrities and 80 non-celebrities. We paired the faces in the collection and used Qwen2.5-VL~\cite{bai2025qwen2} to generate a prompt for each pair, resulting in 12,720 data points. We randomly selected 1,000 of these as our benchmark, and the remaining data is used as reinforcement learning training data.
For multi-object customization, we use the public MUSAR-Gen~\cite{guo2025musar} dataset as our foundation. 
We use Florence-2 and SAM to obtain the detailed positions of reference objects. To ensure segmentation quality, we further employ Qwen2.5-VL to filter the results. 
We split 1,000 samples from this dataset that are not seen during training as a benchmark.


\noindent\textbf{Evaluation metrics.}
Following the ~\cite{le2025one,mou2025dreamo}, we evaluate generated image quality using standard metrics.
We calculate the cosine similarity between the CLIP~\cite{radford2021learning} embeddings of the prompt and the image (CLIP-T) to evaluate text fidelity.
We also use HPSv2 for aesthetic scoring.
For subject fidelity, we employ cosine similarity measures between generated images and reference images within CLIP and DINO~\cite{zhang2022dino} spaces, referred to as CLIP-I and DINO-I scores, respectively.
Note that to accurately calculate CLIP-I and DINO-I, we first use a face detector~\cite{deng2019arcface} or Florence-2 to accurately locate the position of subjects in the generated image, and then calculate them.
Additionally, for multi-human generation, we incorporate Face Similarity (Face-Sim)~\cite{deng2019arcface} and the Hungarian algorithm, enabling them to assess subject fidelity more accurately.

\begin{figure*}[tb]
    \centering
    \includegraphics[width=1.0\linewidth]{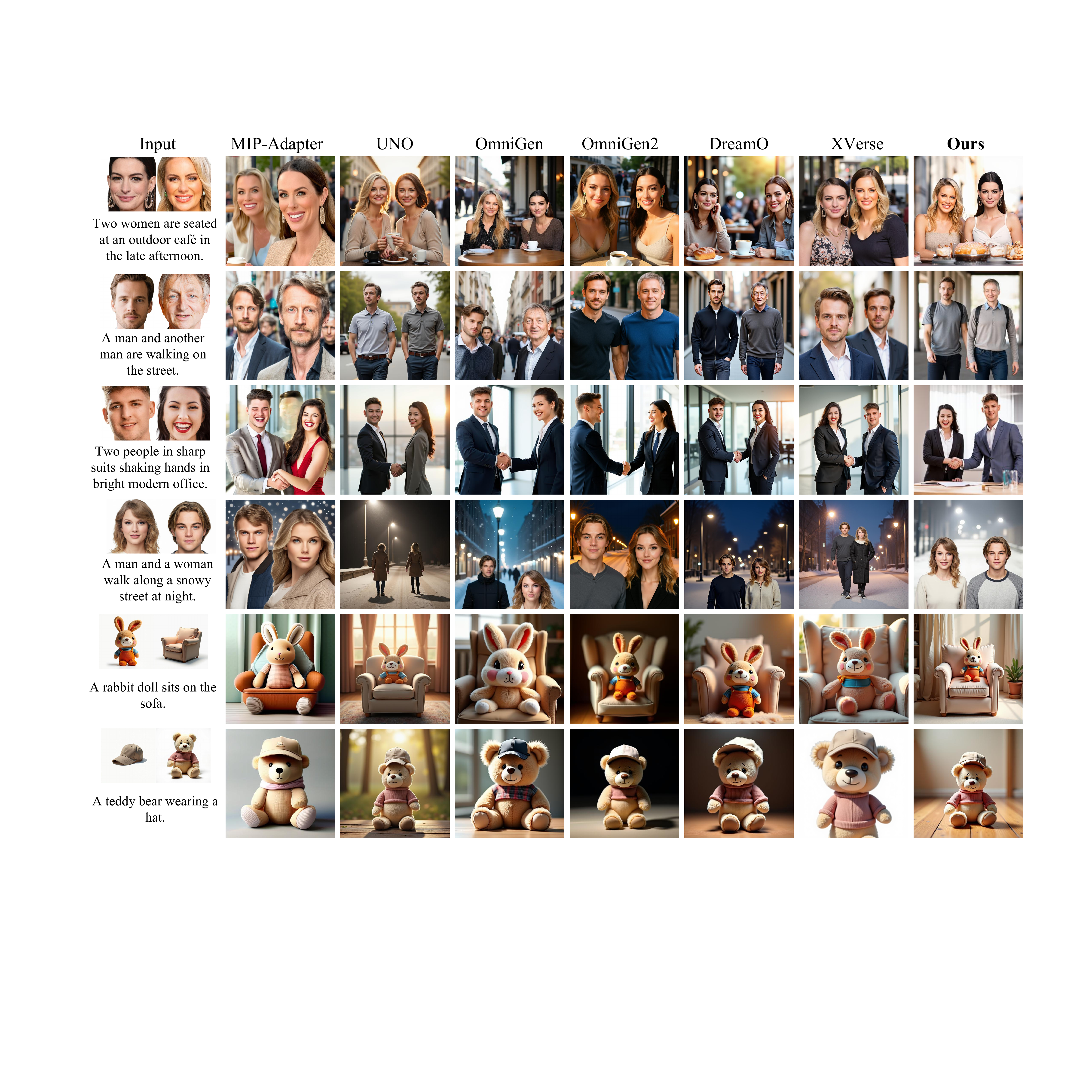}
    \caption{Qualitative comparison with existing methods on the multi-subject generation (Zoom in for best visual comparison). Our method significantly improves subject fidelity while maintaining good text alignment and aesthetics.}
    \label{fig:vis_cmp}
    \vspace{-0.5cm}
\end{figure*}

\subsection{Quantitative Comparison}
%

We conducted comprehensive comparisons with existing methods in both multi-human and multi-object customized generation. 
These methods include MS-Diffusion~\cite{wang2025msdiffusion}, MIP-Adapter~\cite{zhong2025mod}, OmniGen~\cite{xiao2025omnigen}, UNO~\cite{wu2025less}, OmniGen2~\cite{wu2025omnigen2}, DreamO~\cite{mou2025dreamo}, and XVerse~\cite{chen2025xverse}. 
The results, as shown in~\Cref{tab:sota}, demonstrate that our method achieves significant improvements over existing approaches, achieving the highest overall score.
In the multi-human generation task, our model shows a commanding lead in subject fidelity, achieving the top score. 
Notably, the significant 28.3\% relative improvement in the Face-Sim over the next-best method highlights our model's ability to effectively distinguish between different subjects and preserve their individual detailed features. 
This is crucial for accurately and reliably accomplishing the highly sensitive and challenging task of multi-human customization.
Concurrently, the model remains highly competitive in text-image alignment.
This outstanding performance extends to the multi-object generation benchmark, where our method once again achieved top ranks in text alignment and thematic fidelity, and second place in aesthetics, validating its robustness and versatility.
This demonstrates the effectiveness of our decoupling framework, which achieves a strong balance between subject fidelity and multi-dimensional human preferences.
It's worth noting that the decline in aesthetic metrics for multi-person customized generation is due to the quality of the training data. 
We've provided further details in Appendix.
Competitive results achieved on high-quality, open-source, multi-object custom data further support this.

\subsection{Qualitative Comparison}
To better demonstrate the effectiveness of our method, we focused on the more challenging task of multi-human customized generation and conducted a qualitative comparison with existing methods.. 
As shown in the first two rows of \Cref{fig:vis_cmp}, methods trained directly with In-Context Learning, such as UNO, OmniGen, and OmniGen2, struggle with attribute confusion when generating subjects of the same gender, which degrades subject fidelity. 
In contrast, our method accurately preserves the unique features of each individual while maintaining high text alignment and image quality.
Our method maintains strong subject fidelity even in interactive scenarios, as shown in the third row of \Cref{fig:vis_cmp}.
This result validates the effectiveness of our framework. 
For multi-objects generation, as shown in the last two lines of \Cref{fig:vis_cmp}, our method also demonstrates good subject fidelity and generated image quality.
Appendix provides further qualitative comparisons of multi-subject generation (including human and objects) and includes a User Study to demonstrate that our method captures more human preferences.

\subsection{Ablation Studies}
\input{tables/abalation}
\begin{figure}[tb]
    \centering
    \includegraphics[width=0.92\linewidth]{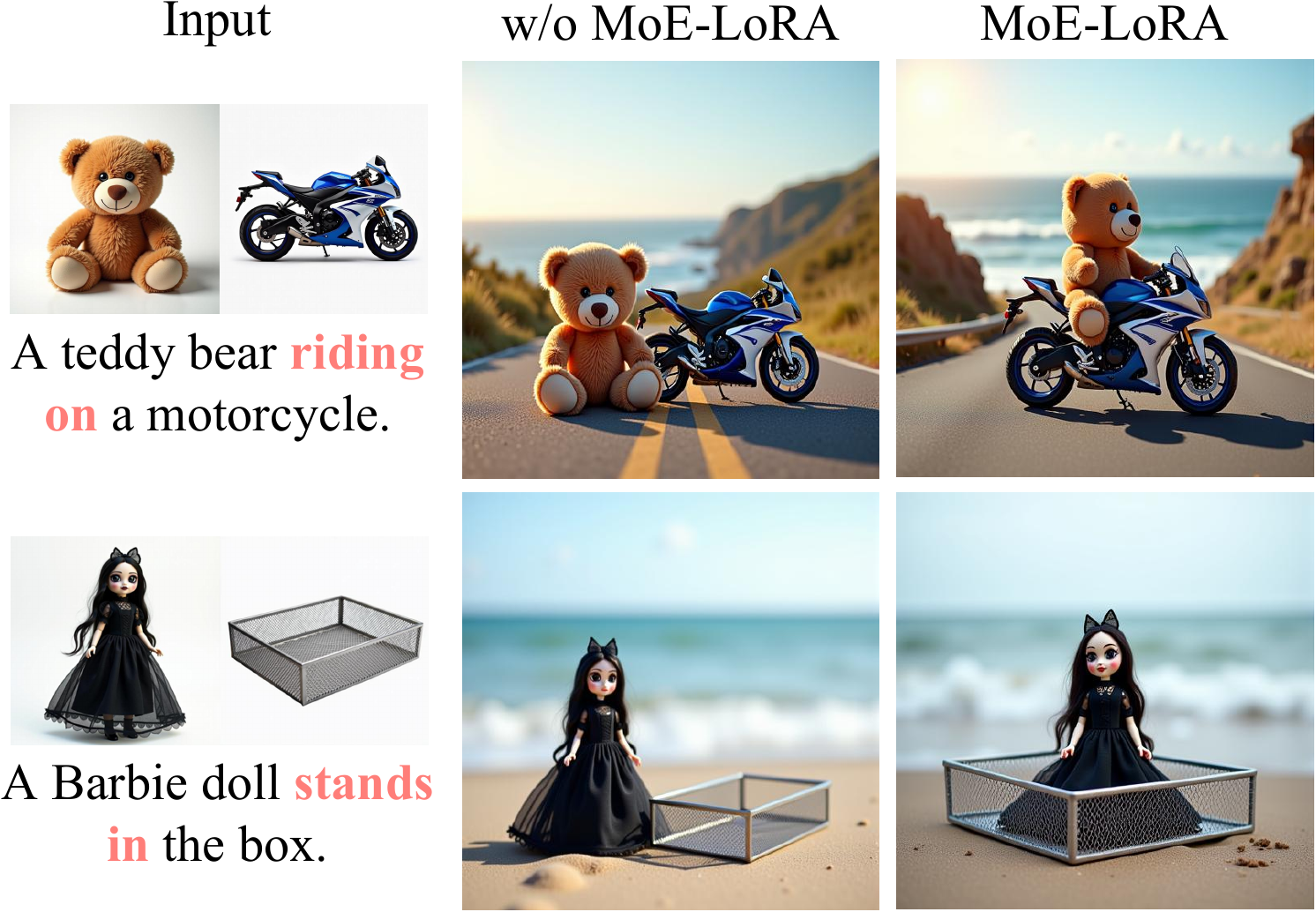}
    \caption{Ablation study on the effectiveness of MoE-LoRA. MoE-LoRA expands the model capacity, enabling the model to better grasp the diversity of spatial layout.}
    \label{fig:moe}
    \vspace{-0.5cm}
\end{figure}

We conduct detailed ablation studies to validate the effectiveness of each component of our decoupling framework. 
Since the multi-human customization evaluation metrics are more accurate, our quantitative ablation experiment is mainly based on multi-human customization. 
We use UNO~\cite{wu2025less}, a Flux-based model trained with highly coupled objectives, as our baseline for comparison, and the detailed results are presented in \Cref{tab:abalation} and \Cref{fig:moe}.

\noindent\textbf{Effect of the Identity-Disentangled Attention Regularization.}
Identity-Disentangled Attention Regularization aims to achieve high-level subject fidelity in post-training.
As shown in \Cref{tab:abalation}, the base model performs poorly in terms of subject fidelity, achieving a Face-Sim score of only 0.1474. 
Upon introducing our $\mathcal{L}_{attn}$ using single LoRA, we observe a dramatic improvement across all subject fidelity metrics.
As shown on the left of \Cref{fig:intro}, our method successfully separates the attention of different subjects.
However, it can be seen that $\mathcal{L}_{attn}$ leads to a decrease in text alignment.
As shown in \Cref{fig:moe}, this inability of a single LoRA to learn complex spatial layouts caused by different prompts and subjects results in some spatial layouts failing to be generated correctly, \eg, unable to generate the scene of a teddy bear riding on a motorcycle.

\noindent Building on $\mathcal{L}_{attn}$, we integrate the MoE-LoRA architecture to expand model capacity, which yields further gains across subject fidelity while also improving text alignment. 
Furthermore, as shown in \Cref{fig:moe}, the combinatorial samples that could not be generated correctly before have been significantly improved.
This demonstrates that utilizing MoE-LoRA to expand model capacity can effectively alleviate the challenges of diverse spatial layouts.

\noindent\textbf{Effect of Identity-Preserving Preference Optimization.}
We introduce IPPO to align human preferences and ensure subject fidelity during post-training.
As shown in ~\Cref{tab:abalation}, the introduction of IPPO significantly improves text alignment and aesthetic metrics, while further enhancing subject fidelity, demonstrating the effectiveness of this method.

%% file: tables/sota.tex
\begin{table*}[]
    \centering
    \resizebox{0.96\linewidth}{!}{
    \begin{tabular}{lcccccccccc}
        \toprule
        \multirow{2}{*}{Methods} & \multicolumn{4}{c}{Multi Human Genertaion}                            & \multicolumn{5}{c}{Multi Object Generation}                                            & Overall         \\ \cmidrule(l{3pt}r{3pt}){2-6}  \cmidrule(l{3pt}r{3pt}){7-10} \cmidrule(l{3pt}r{3pt}){11-11}
                                 & CLIP-T          & Face-Sim        & DINO-I          & CLIP-I          & AES             & CLIP-T          & DINO-I          & CLIP-I          & AES             & AVG             \\\midrule
        MS-Diffusion             & 0.2498          & 0.0945          & 0.4767          & 0.5801          & 0.2461          & 0.2887          & 0.4002          & 0.6681          & 0.2685          & 0.3636          \\
MIP-Adapter              & 0.2631          & 0.2117          & 0.6959          & 0.7140          & 0.2791          & 0.2984          & 0.5470          & 0.7776          & 0.2374          & 0.4471          \\
OmniGen                  & 0.2741          & 0.3238          & 0.7225          & 0.7642          & 0.2996          & 0.3151          & 0.6727          & 0.8085          & 0.2561          & 0.4930          \\
UNO                      & 0.2645          & 0.1474          & 0.5972          & 0.6489          & 0.2954          & 0.3259          & 0.7374          & 0.8392          & 0.2676          & 0.4582          \\
OmniGen2                 & \textbf{0.2837} & 0.2453          & 0.6788          & 0.7205          & \textbf{0.3103} & \underline{0.3310}    & \underline{0.7538}    & \underline{0.8470}    & \textbf{0.2872} & 0.4953          \\
DreamO                   & 0.2747          & 0.3345          & 0.7441          & 0.7988          & \underline{0.3056}    & 0.3207          & 0.7394          & 0.8393          & 0.2637          & 0.5134          \\
XVerse                   & 0.2591          & \underline{0.4117}    & \underline{0.7665}    & \underline{0.8027}    & 0.2498          & 0.2981          & 0.7449          & 0.8456          & 0.2595          & \underline{0.5153}    \\
\rowcolor{blue!10}
Ours                     & \underline{0.2753}    & \textbf{0.5284} & \textbf{0.8294} & \textbf{0.8524} & 0.2915          & \textbf{0.3380} & \textbf{0.7824} & \textbf{0.8608} & \underline{0.2748}    & \textbf{0.5592}  \\ \bottomrule
        \end{tabular}}
    \caption{Quantitative comparison with SoTA methods on the multi-human and multi-object generation. The best results are in \textbf{bold}, and the second-best are \underline{underlined}. Our method outperforms others, especially in subject fidelity, and achieves the highest overall score.}
    \label{tab:sota}
    \vspace{-0.1cm}
\end{table*}

%% file: tables/abalation.tex
\begin{table}[]

    \centering

    \resizebox{1.0\linewidth}{!}{

    \begin{tabular}{ccccccccc}

    \toprule

    $\mathcal{L}_{attn}$ & MoE-LoRA & IPPO & CLIP-T & Face-Sim & DINO-I & CLIP-I & AES    & Overall \\ \midrule

    \ding{55}     & \ding{55}       &\ding{55}& 0.2645 & 0.1474   & 0.5972 & 0.6489 & 0.2954 & 0.3907 \\

    \checkmark    & \ding{55}         &\ding{55}      & 0.2637 & \cellcolor{blue!5}0.4983   & \cellcolor{blue!5}0.7953 & \cellcolor{blue!5}0.8032 & 0.2653 & 0.5252 \\

    \checkmark    & \checkmark        & \ding{55}     & 0.2674 & \cellcolor{blue!15}0.5154   & \cellcolor{blue!15}0.8107 & \cellcolor{blue!15}0.8480  & 0.2661 & 0.5415 \\

    \checkmark    & \checkmark        & \checkmark    & 0.2753 & \cellcolor{blue!25}0.5284   & \cellcolor{blue!25}0.8294 & \cellcolor{blue!25}0.8524 & 0.2915 & 0.5554 \\ \bottomrule

    \end{tabular}

    }

    \caption{Ablation study of our methods. $\mathcal{L}_{attn}$ is \Cref{eq:dice_loss} in Identity-Disentangled Attention Regularization, and IPPO is our Identity-Preserving Preference Optimization. The results demonstrate the effectiveness of each module.}

    \label{tab:abalation}

    \vspace{-0.2cm}

\end{table}

%% file: sections/6_conclusion.tex
\section{Conclusion}
\label{sec:conclusion}
In this paper, we address the coupled training objective of existing multi-subject generation methods, where a single loss fails to optimize for both subject fidelity and human preference.
We introduce a framework that decouples this problem into two stages.
First, a pre-training stage uses explicit positional supervision to fix attention bleeding and achieve high fidelity.
Second, a post-training stage employs an identity-preserving reinforcement learning framework to align with human preferences while maintaining this fidelity.
Our decoupled approach achieves state-of-the-art results in both fidelity and preference alignment, validating its effectiveness.

%% file: sections/X_appendix.tex
\clearpage
\maketitlesupplementary
\appendix
\renewcommand{\thesection}{\Alph{section}}
\renewcommand{\thesubsection}{\Alph{section}.\arabic{subsection}}
\setcounter{page}{1}
\setcounter{section}{0}
\setcounter{figure}{0}
\setcounter{table}{0}

\section{More Details for Identity-Preserving Preference Optimization}
\label{sec:app_ippa}
We provide a detailed algorithm execution flow for Identity-Preserving Preference Optimization~\Cref{sec:rl}, as shown in \Cref{algo:mixgspo}.
\begin{algorithm*}[h]
\caption{Identity-Preserving Preference Optimization Training Process}
\label{algo:mixgspo}
\small
\begin{algorithmic}[1]
\Require initial policy model $\pi_\theta$; composite reward model $R$; prompt dataset $\mathcal{C}$; reference subjects dataset $\mathcal{Z}_{\text{data}}$; total sampling steps $T$; number of samples per prompt $N$; sliding window $W(l)$, window size $w$, shift interval $\tau$, window stride $s$
\State Init left boundary of $W(l)$: $l \gets 0$
\For{training iteration $m=1$ \textbf{to} $M$}
    \State Sample batch prompts $\mathcal{C}_b \sim \mathcal{C}$ and corresponding subjects $\mathcal{Z}_b \sim \mathcal{Z}_{\text{data}}$
    \State Update old policy model: $\pi_{\theta_{\text{old}}} \gets \pi_\theta$
    \For{each prompt $c \in \mathcal{C}_b$ and subject $\mathcal{Z} \in \mathcal{Z}_b$}
        \State Init the same noise $s_0 \sim \mathcal{N}(0,\mathbf{I})$
        \For{generate $i$-th image from $i=1$ \textbf{to} N}
            \For{sampling timestep $t=0$ \textbf{to} $T-1$} 
            \Comment{$\pi_{\theta_{\text{old}}}$ mixed sampling loop}
                \If{$t \in W(l)$}
                    \State Use SDE Sampling to get $s^{i}_{t+1}$ 
                \Else
                    \State Use ODE Sampling to get $s^{i}_{t+1}$
                \EndIf
            \EndFor
        \EndFor
        \State Calculate advantage: $A_i \gets \frac{R(s^i_T, c, \mathcal{Z}) - \text{mean}(\{R(s^j_T, c, \mathcal{Z})\}_{j=1}^N)}{\text{std}(\{R(s^j_T, c, \mathcal{Z})\}_{j=1}^N)}$
        \For{optimization timestep $t \in W(l)$} \Comment{optimize policy model $\pi_\theta$}
            \State Update policy model via gradient ascent: $\theta \gets \theta + \eta\nabla_\theta\mathcal{J}_{GSPO}$
        \EndFor
    \EndFor
    \If{$m \bmod \tau = 0$} \Comment{move sliding window}
        \State $l \gets \min(l+s,~T-w)$ 
    \EndIf
\EndFor
\end{algorithmic}
\end{algorithm*}

\section{More Implementation Details.}
\label{sec:app_imp_details}
In this section, we provide more details on the hyperparameter settings and specific training details. 
As mentioned in the main text, we decouple the training process into two parts, and the details of each phase are as follows:

\noindent\textbf{Subject Fidelity Focused Pre-training.} 
As described in \Cref{sec:over}, in this stage, we train the model to primarily ensure it can generate images with high subject fidelity.
Following~\cite{wu2025less,chen2025xverse}, we adopt a progressive training approach, first pre-training on single-subject data, and then using this as a foundation to train on multi-subject data.
\textit{1. Single-Subject Pre-training.} We first pre-train the model for 40,000 steps on an internal single-subject dataset to equip it with foundational subject customization capabilities. In this stage, only $\mathcal{L}_{diff}$ is used for supervision. We use the AdamW optimizer with a $3 \times 10^{-5}$ learning rate and a weight decay of $1 \times 10^{-2}$. We use 8 cards for training and set the batch size of each card to 6.
\textit{2. Multi-Subject Customization Training.} Then we train two models for customized human generation and object generation, respectively. 
For multi-human customized generation, we decrease the learning rate to $1 \times 10^{-5}$, set the loss weight $\lambda=0.3$, and introduce the attention loss $\mathcal{L}_{attn}$. This stage runs for 25,000 steps. We use 8 cards at this stage and set the batch size to 4.
For multi-object customized generation, since the size of the dataset is smaller than multi-human datasets, we only train for 15,000 steps.

\noindent\textbf{Multi-Dimensional Preference Alignment Post-training.} 
As described in \Cref{sec:over}, after completing the Subject Fidelity Focused Pre-training, in the second stage, we primarily use post-training to align it with multi-dimensional human preferences.
We fine-tune the model using our proposed Identity-Preserving Preference Optimization. Following~\cite{li2025mixgrpo}, we configure with a sampling step of 16, a window size of $w=2$, a shift interval of $\tau=50$, and a window stride of $s=1$. This stage consists of 300 steps. For multi-human customization, the reward weights are set to $w_{id}=0.5$, $w_{text}=1.4$, and $w_{aes}=0.7$. For multi-object customization, we adjust the subject fidelity weight to $w_{id}=1.0$, while keeping $w_{text}=1.4$ and $w_{aes}=0.7$.
In this stage, we used 16 cards for training and set the batch size of each card to 1.

\section{Training Dataset Construction Pipeline.}
\label{sec:app_train_data}

\begin{figure*}[htb]
    \centering
    \includegraphics[width=0.98\linewidth]{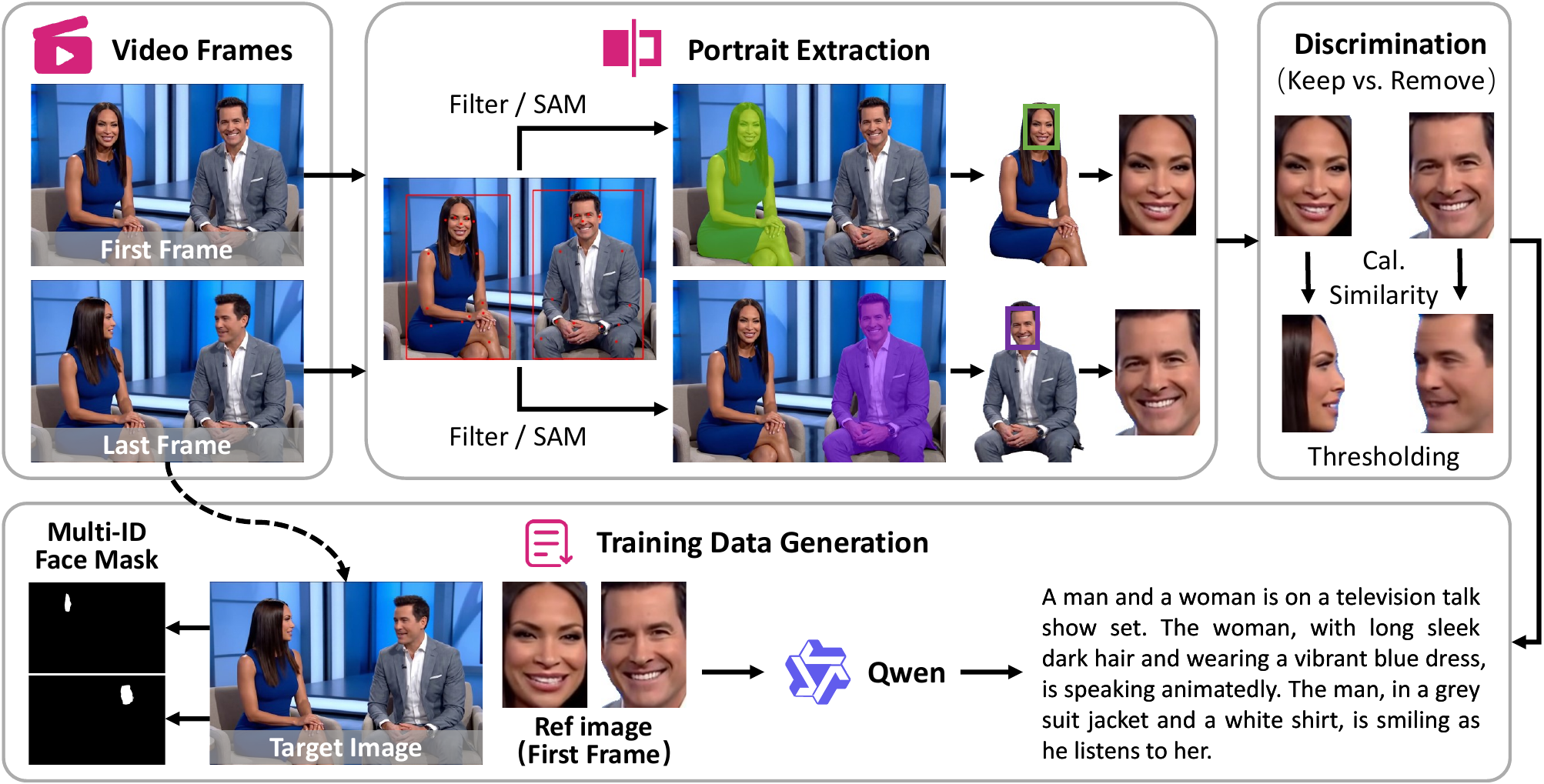}
    \caption{Data processing pipeline for customized multi-human image generation. }
    \label{fig:datapipe}
\end{figure*}

Due to the scarcity of public multi-human customization datasets with adequate annotations, we designed a comprehensive and automated data preparation process to extract training data from OpenHumanVid~\cite{li2025openhumanvid} video datasets, as shown in~\Cref{fig:datapipe}. 
This pipeline processes raw video clips to generate structured training samples, each containing a target image with multiple subjects, corresponding identity reference images, segmentation masks, and a detailed textual description. The entire workflow ensures subject fidelity, high image quality, and rich annotation. Specifically, we sample video clips featuring two individuals from a large database. The process begins with frame selection and subject localization. For each input video, we sample an initial frame as the source for reference images and a middle frame as the target scene. We first employ a YOLO-Pose~\cite{maji2022yolo} to obtain initial bounding boxes and keypoint information for each person. Following localization, we leverage the Segment Anything Model (SAM)~\cite{ravi2024sam} to generate high-fidelity segmentation masks for each individual, effectively isolating them from the background. To refine this output, only the largest connected component of the mask is retained.
A critical subsequent step is ensuring subject fidelity across frames. We apply a face detection model~\cite{deng2019arcface} to the segmented portraits to locate facial regions, and then use a face recognition model~\cite{deng2019arcface} to extract a normalized feature embedding for each face. By computing the cosine similarity between embeddings from the reference and target frames, we enforce a stringent threshold to discard pairs where identity cannot be confidently verified.
Once a pair of frames passes this verification, the pipeline generates the final training sample. The segmented portraits and cropped faces from the initial frame are saved as the reference =images. The target frame is cropped into a square as target image, with its corresponding body and face masks preserved. Finally, a powerful vision-language model, Qwen2.5-VL~\cite{bai2025qwen2}, is prompted with the reference images and the target image to produce a rich text prompt of the entire scene, ensuring descriptive consistency for each subject.

\input{tables/data_analysis}
\begin{figure}[]
    \centering
    \includegraphics[width=1.0\linewidth]{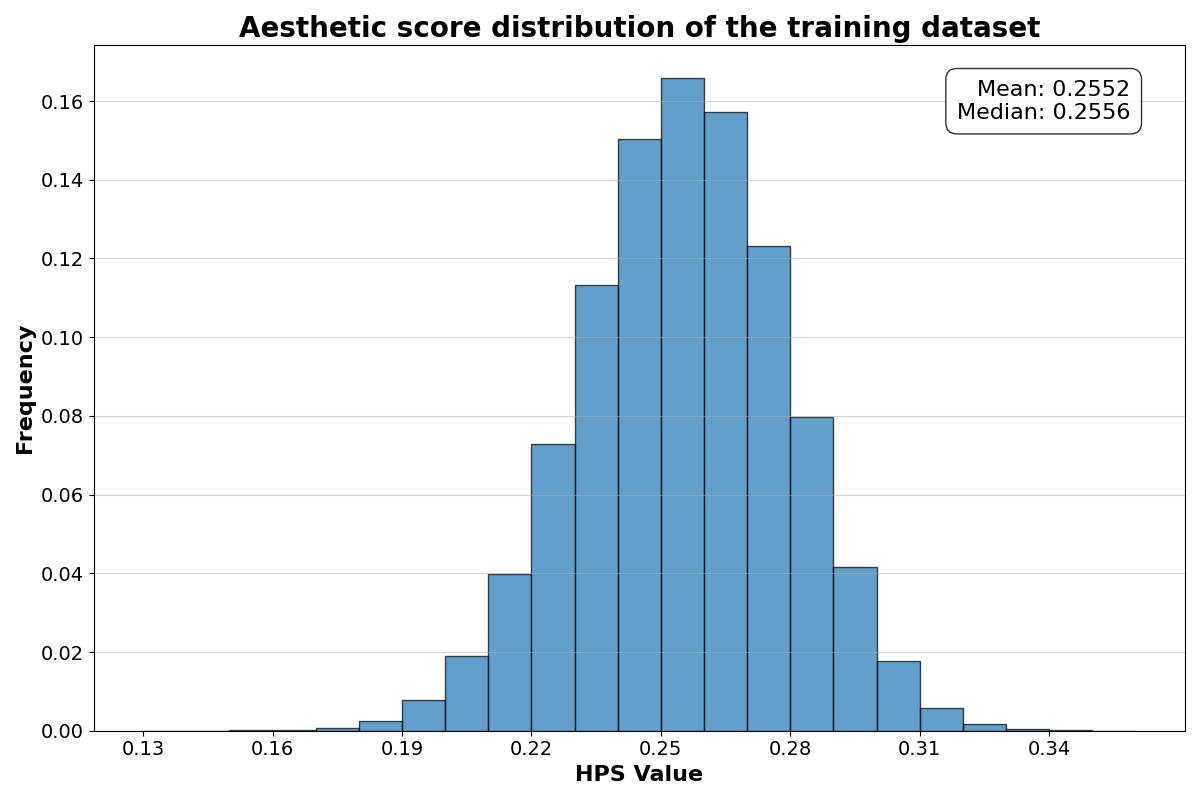}
    \caption{Aesthetic Score Distribution of Training Data. The histogram illustrates the frequency of HPS v2 scores within our training dataset. The distribution is centered around a mean of 0.2552.}
    \label{fig:hps_data}
\end{figure}

\section{Impact Analysis of Dataset Quality}
\label{sec:data_qulity}
To investigate whether the variance in aesthetic metrics between our method and certain baselines for multi-human generation task (e.g., UNO~\cite{wu2025less}) stems from algorithmic limitations or discrepancies in training data quality, we conducted a detailed statistical analysis and ablation study. It is important to note that many existing methods utilize private, high-quality datasets, whereas our model is trained on a subset curated from open-source video data, which inevitably contains frames with motion blur or lower aesthetic appeal.

\noindent\textbf{Data Aesthetic Distribution.} 
We first analyzed the aesthetic quality of our training dataset using the HPS v2 scoring model. As illustrated in Fig.~\ref{fig:benchmark_samples}, the aesthetic scores of our training samples follow a normal distribution with a mean of $0.2552$ and a median of $0.2556$. This relatively low baseline suggests that the model's "upper bound" for aesthetics is naturally constrained by the training data when using standard supervised learning.

\noindent\textbf{Baseline Retraining and Comparison.}
To verify this hypothesis, we conducted ablation study using UNO~\cite{wu2025less} as a baseline. We retrained the UNO model on our dataset (denoted as UNO$^\dag$) and compared it with the official UNO release and our method.
As shown in Tab.~\ref{tab:data_ablation}, when UNO is retrained on our data, its aesthetic score (AES) drops significantly from $0.2954$ (Official) to $0.2656$. Crucially, this result ($0.2656$) is highly consistent with the performance of our method before the post-training stage (Ours w/o IPPO, $0.2661$). This observation supports two key conclusions:
\begin{itemize}
    \item The decline in aesthetic quality is primarily attributable to the training data rather than the model architecture.
    \item Even with suboptimal data, our full method (Ours Full) successfully recovers the aesthetic quality to $0.2915$ via the proposed Identity-Preserving Preference Optimization (IPPO).
\end{itemize}
This demonstrates that our two-stage decoupled framework effectively aligns the model with human aesthetic preferences, overcoming the limitations of the underlying training data quality.

\section{More Details for Our Benchmark.}
\label{sec:app_bench}

\begin{figure*}[]
    \centering
    \includegraphics[width=1.0\linewidth]{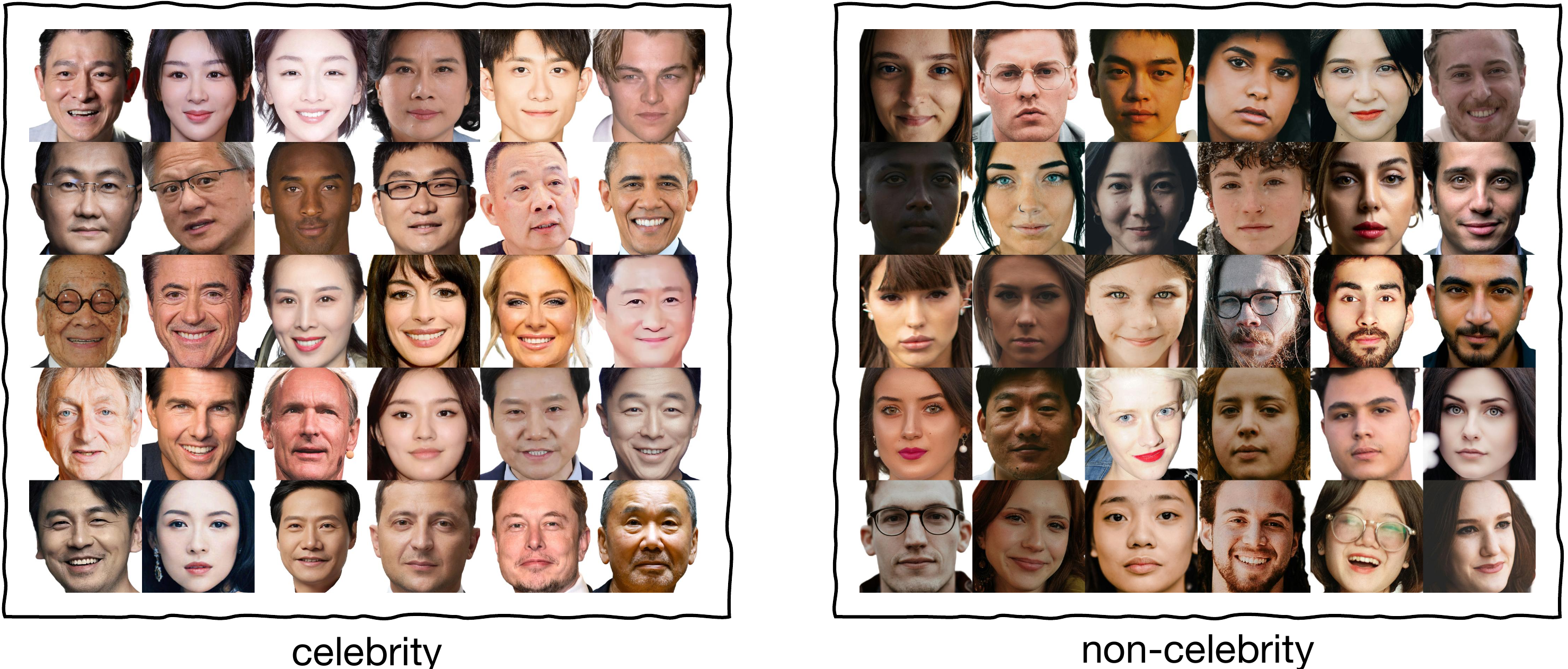}
    \caption{Visualization for part of our multi-human evaluation benchmarks. }
    \label{fig:benchmark_samples}
\end{figure*}

To advance research on high-fidelity multi-subject generation, we constructed a benchmark dataset by collecting images from publicly available sources and extracting the corresponding facial regions as reference images. 
The dataset comprises 80 celebrities and 80 non-celebrities, covering diverse attributes in terms of gender, age, and ethnicity (male/female; young/elderly; Caucasian, Black, and Asian). 
We used this face collection as a reference pool and paired faces within it. For each pair, we employed Qwen2.5-VL to generate distinctive natural-language prompts to provide diverse textual descriptions. 
Representative samples are shown in \Cref{fig:benchmark_samples}. 

\section{Limitation.}
\label{sec:app_limit}
Although MultiCrafter has achieved excellent performance in multi-subject driven image generation tasks, our work still has certain limitations, which also point the way for future research. 

First, the scale and quality of the training data are the primary limiting factors. 
We discussed in detail the limitations imposed by the quality of our data in ~\Cref{sec:data_qulity}.
Currently, high-quality, publicly available datasets for multi-subject driven generation remain scarce~\cite{wu2025less,chen2025xverse}. Although we have designed a complete automated data processing pipeline to extract training samples from videos, our dataset is still limited in scale and diversity due to the quantity and quality of open-source video data. 

Second, the effectiveness of our method has been primarily validated in two-subject scenarios.
Since our multi-person dataset and the public MUSAR dataset~\cite{guo2025musar} mainly contain two subjects, the experiments in this paper were centered around this setting. Although we have supplemented some results from a small number of customized generation experiments with 3 person in ~\Cref{sec:app_multi_human}, the model’s generation capabilities when dealing with more objects have not yet been fully validated.
It is worth noting that our framework was designed with scalability in mind; both the attention regularization mechanism and the Multi-ID Alignment Reward (based on the Hungarian algorithm) in the reinforcement learning framework can be directly extended to scenarios with more subjects. 

For future work, we plan to explore improvements from both data and model perspectives. On one hand, we will attempt to construct larger, higher-quality datasets containing a more diverse number of subjects by combining synthetic data with image editing~\cite{wu2025qwen} techniques. On the other hand, we will train and evaluate the model in scenarios with more subjects to further enhance the generalization and robustness of MultiCrafter, enabling it to handle more complex personalized image generation.


\section{User Study}
\begin{figure}[]
  \centering
  \includegraphics[width=1.0\linewidth]{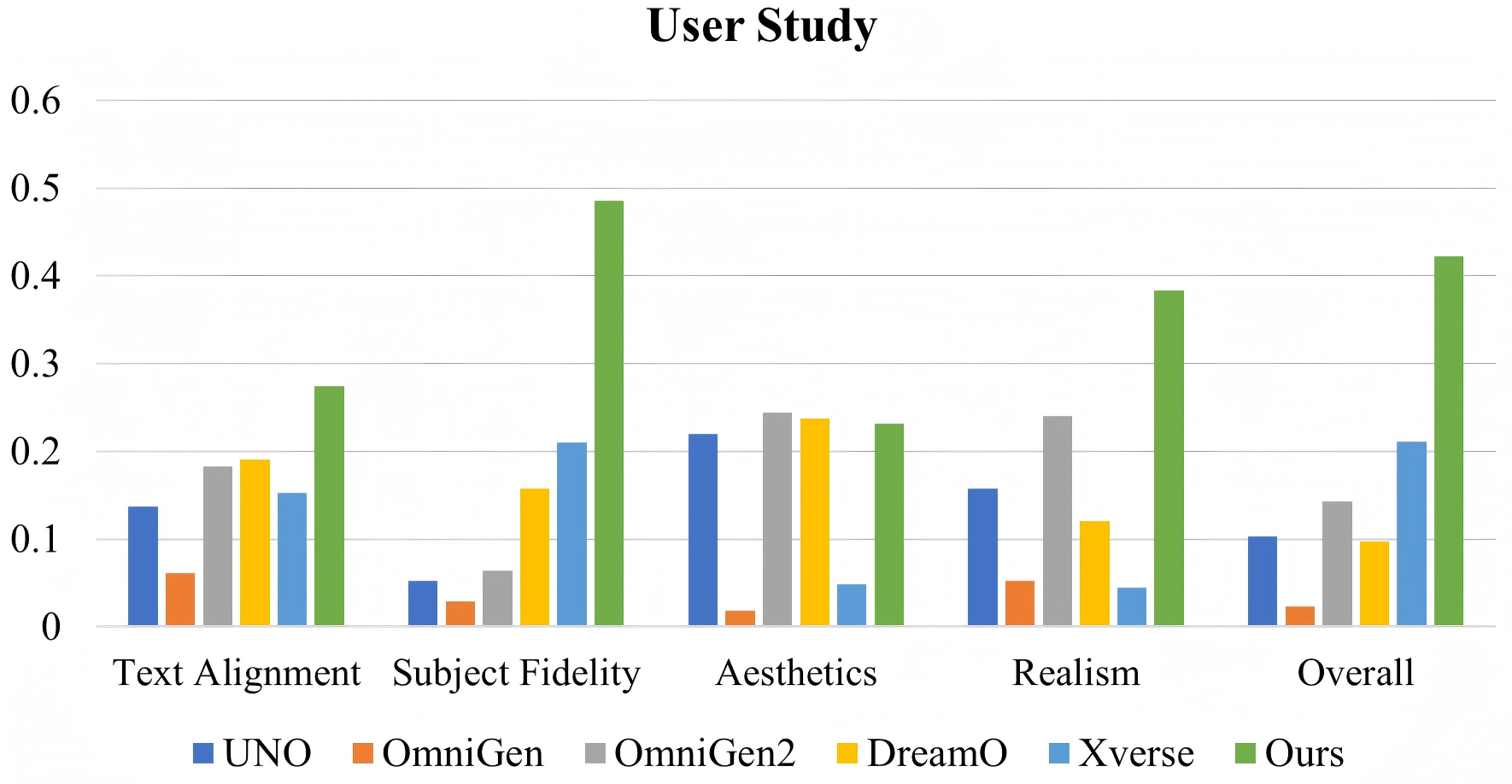} 
  \caption{User Study. Our method achieved the best results in both subject fidelity and overall quality, demonstrating its effectiveness.}
  \label{fig:app_moe_router}
\end{figure}
To complement our quantitative analysis and assess human subjective preference, we conducted a user study comparing MultiCrafter against five state-of-the-art baselines: UNO~\cite{wu2025less}, OmniGen~\cite{xiao2025omnigen}, OmniGen2~\cite{wu2025omnigen2}, DreamO~\cite{mou2025dreamo}, and XVerse~\cite{chen2025xverse}. We evaluated the generated results across five key dimensions: text alignment, subject fidelity, aesthetics, realism, and overall preference. As illustrated in \Cref{fig:user_study}, our method demonstrates a commanding lead in subject fidelity, significantly outperforming the second-best method. This result strongly validates the effectiveness of our Identity-Disentangled Attention Regularization in resolving attribute leakage and preserving intricate identity details. Furthermore, MultiCrafter achieves the highest scores in realism and overall quality, while maintaining a top-tier performance in text alignment and competitive aesthetics. These findings confirm that our decoupled training framework successfully resolves the trade-off between fidelity and preference, producing images that are not only faithful to the user-provided subjects but also aesthetically pleasing and semantically accurate.

\section{More Discussion about MoE-LoRA.}
To further validate the effectiveness of our MoE-LoRA architecture in expanding model capacity and handling diverse spatial layouts, we analyzed the expert routing behavior under distinct spatial instructions. 
If the MoE mechanism truly functions as intended, the routing network should dynamically allocate different experts to handle different spatial layouts.
We designed a controlled experiment using two distinct spatial prompts: a `side-by-side' (left-right) layout and a `top-down' (up-down) layout. To isolate the impact of spatial configuration, we kept all other variables constant, including the subject identities, background descriptions, and random seeds. We conducted inference across 4 different seeds and calculated the average change ratio of the expert routing distribution within the Double Blocks. The change ratio for an expert $e$ at layer $l$ is defined as $|P_{A}^{(l,e)} - P_{B}^{(l,e)}| / (P_{A}^{(l,e)} + \epsilon)$, where $P_{A}$ and $P_{B}$ represent the routing probabilities under the two different layouts.

\begin{figure}[t]
  \centering
  \includegraphics[width=1.0\linewidth]{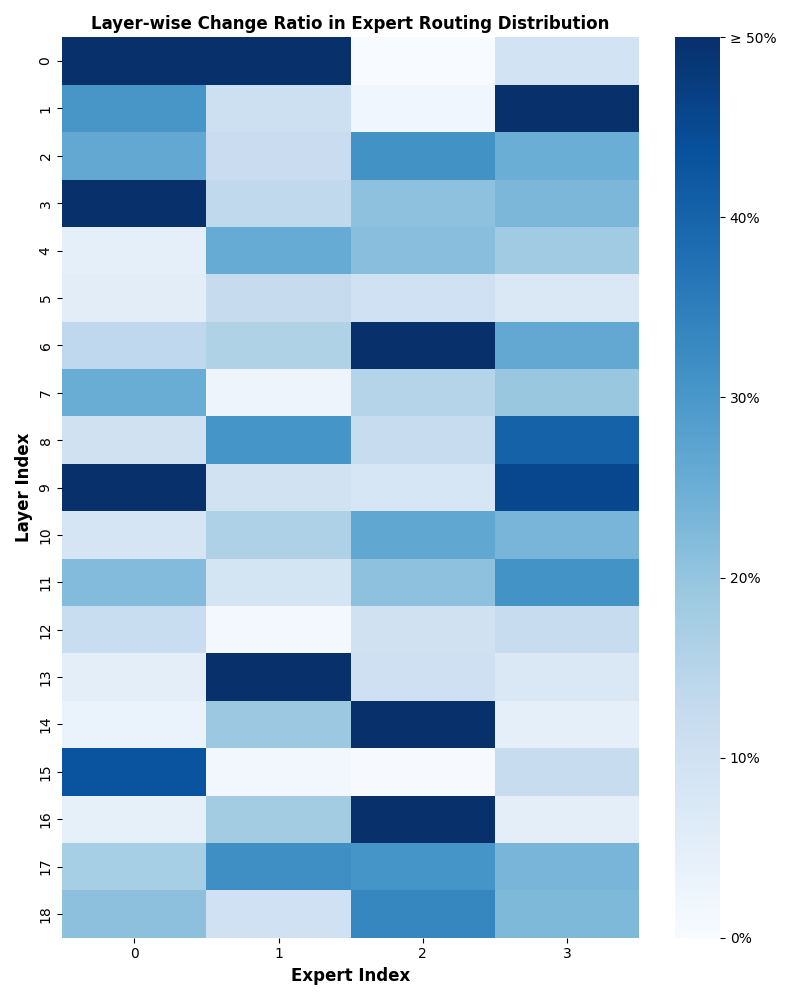} 
  \caption{Layer-wise Change Ratio in Expert Routing Distribution. The heatmap illustrates the difference in expert activation between "Left-Right" and "Top-Down" layouts.}
  \label{fig:app_moe_router}
\end{figure}

\begin{figure*}[t]
  \centering
  \includegraphics[width=1.0\linewidth]{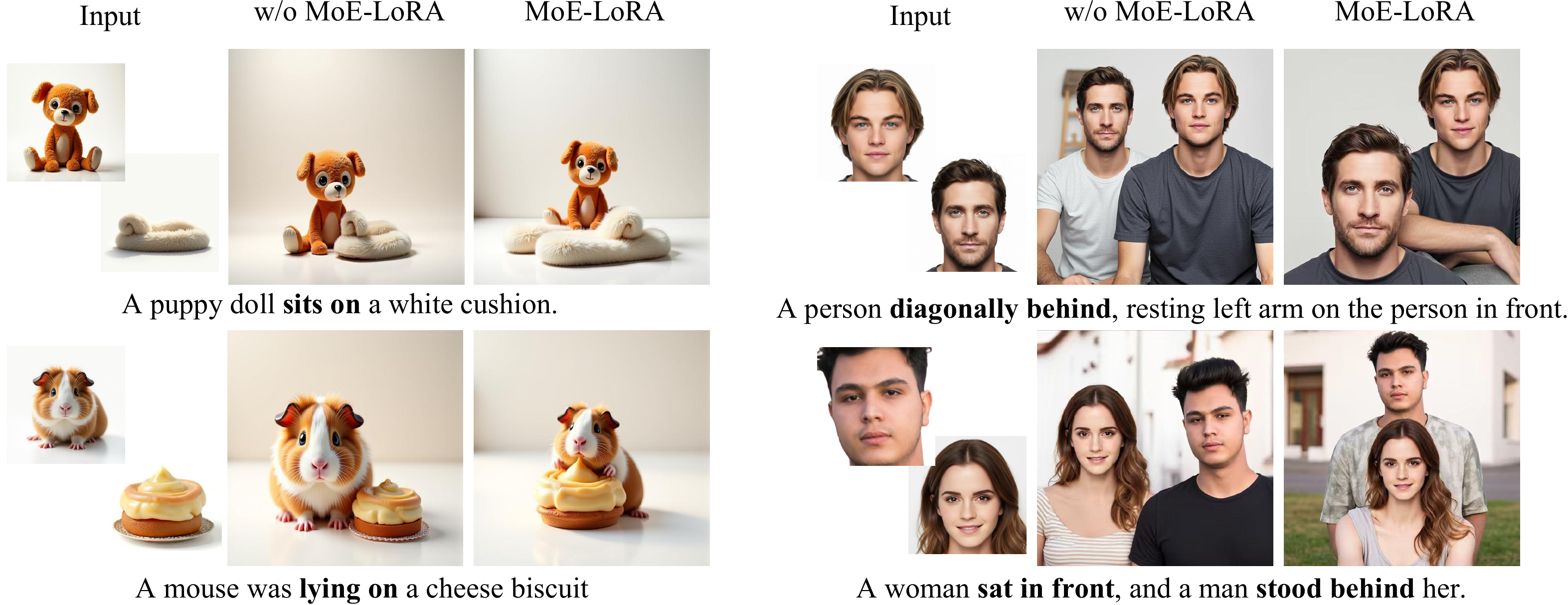} 
  \caption{More Visualization of the effectiveness of MoE-LoRA.}
  \label{fig:app_moe}
\end{figure*}

The results are visualized in \Cref{fig:app_moe_router}. As shown in the heatmap, significant variations in expert activation (indicated by deep blue regions, where the change ratio exceeds $40\%-50\%$) are observed across many layers. 
This phenomenon proves that the routing network effectively perceives the change in spatial prompts. The distinct activation patterns confirm that the model utilizes different experts to construct different layouts.
This experiment strongly supports our claim that MoE-LoRA successfully decouples the learning of diverse spatial scenarios. By assigning specialized experts to different spatial layouts, the framework avoids the "attribute averaging" issue inherent in single-LoRA approaches, thereby achieving high fidelity across complex and varied compositions.
In addition, we have added more visualizations of the results for multi-human and multi-subject generation. As shown in \Cref{fig:app_moe}, after adding MoE-LoRA, many scenes that could not be generated before adding MoE-LoRA can be generated accurately.

\section{More Results of Multi-Human Generation.}
\label{sec:app_multi_human}

\begin{figure*}[tb]
    \centering
    \includegraphics[width=1.0\linewidth]{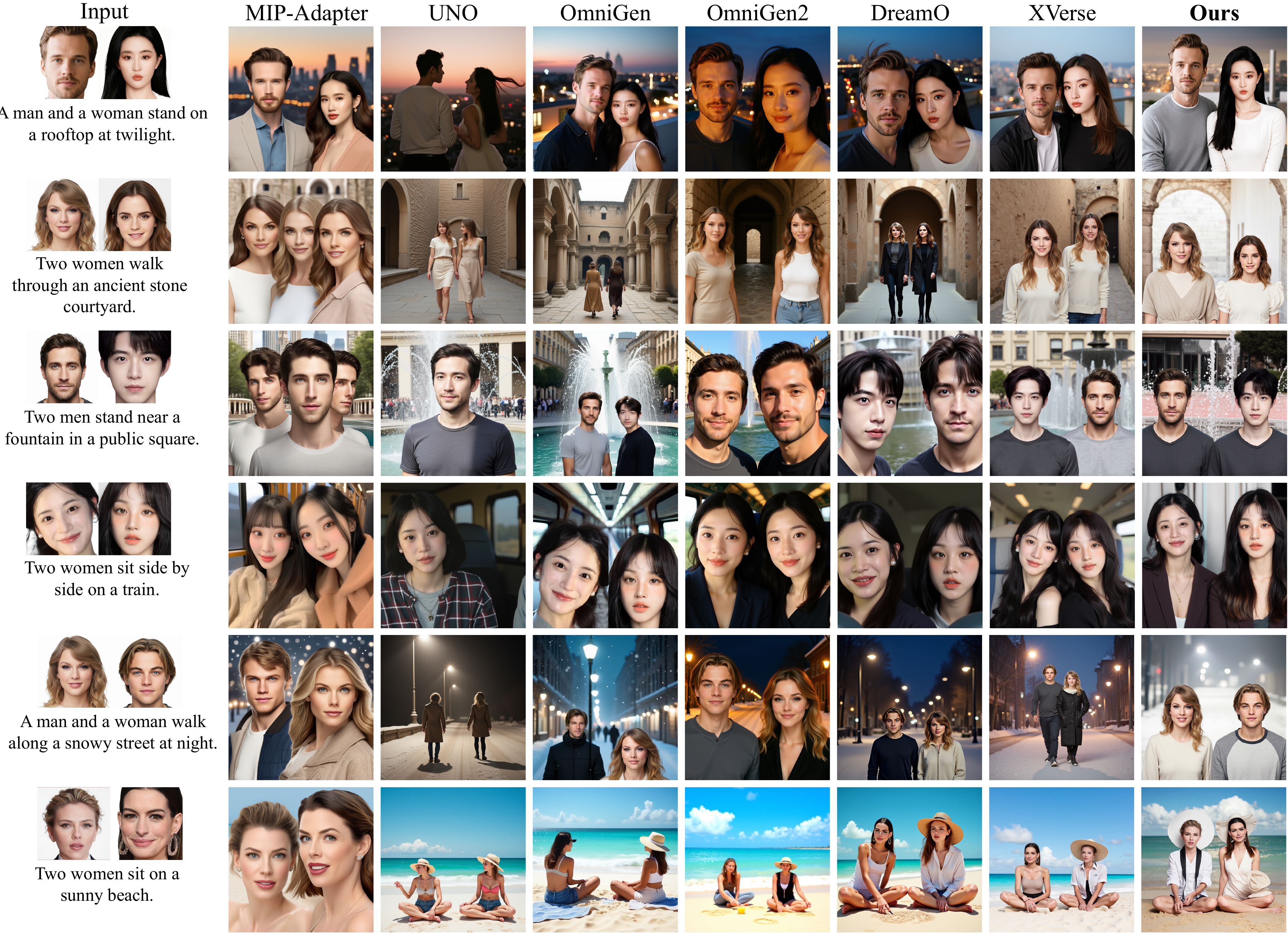}
    \caption{More Visualization of our method in Multi-human Generation.}
    \label{fig:cmp_multi_human1}
\end{figure*}
\begin{figure*}[tb]
    \centering
    \includegraphics[width=1.0\linewidth]{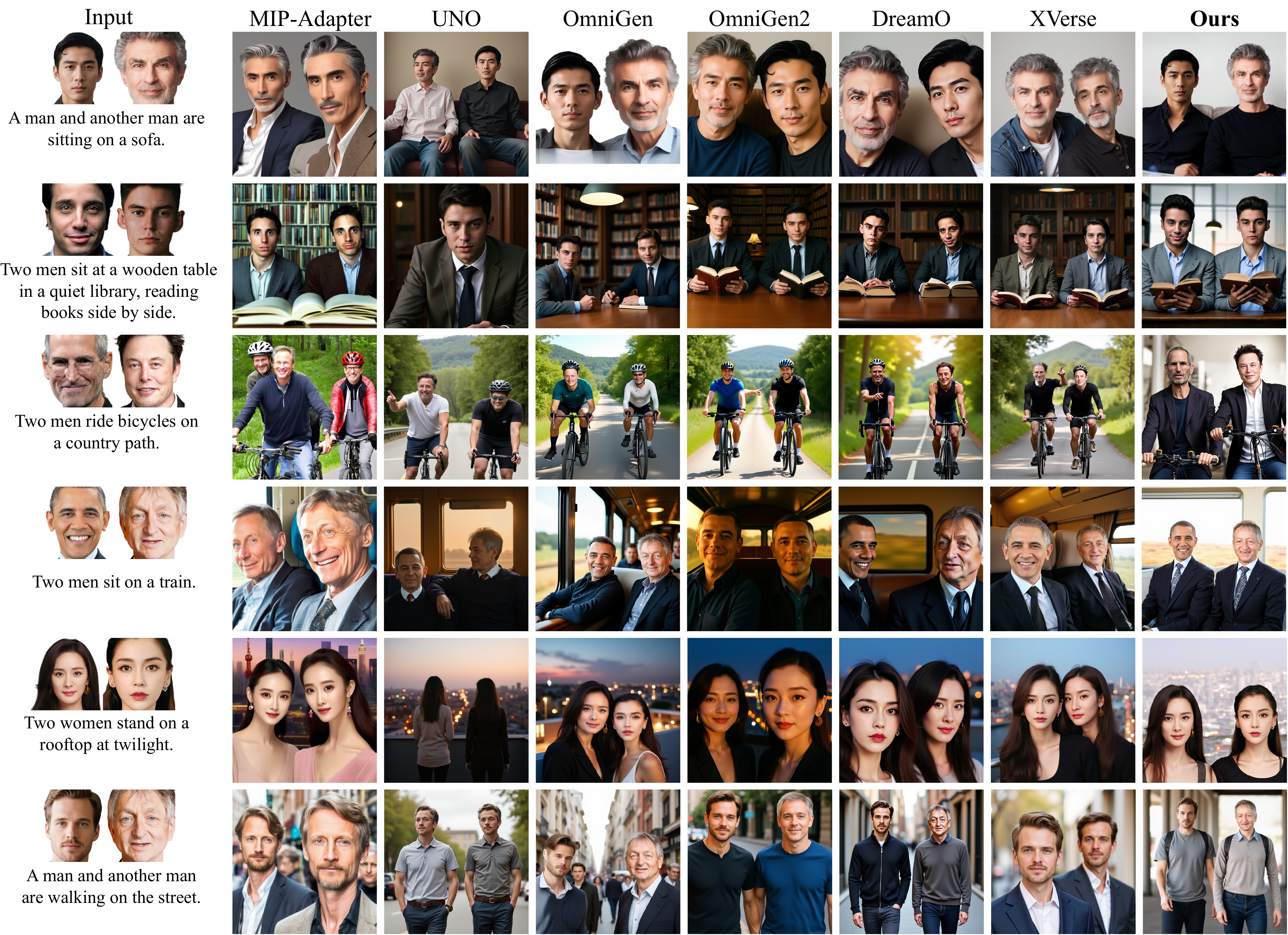}
    \caption{More Visualization of our method in Multi-human Generation.}
    \label{fig:cmp_multi_human2}
\end{figure*}
\begin{figure*}[tb]
    \centering
    \includegraphics[width=1.0\linewidth]{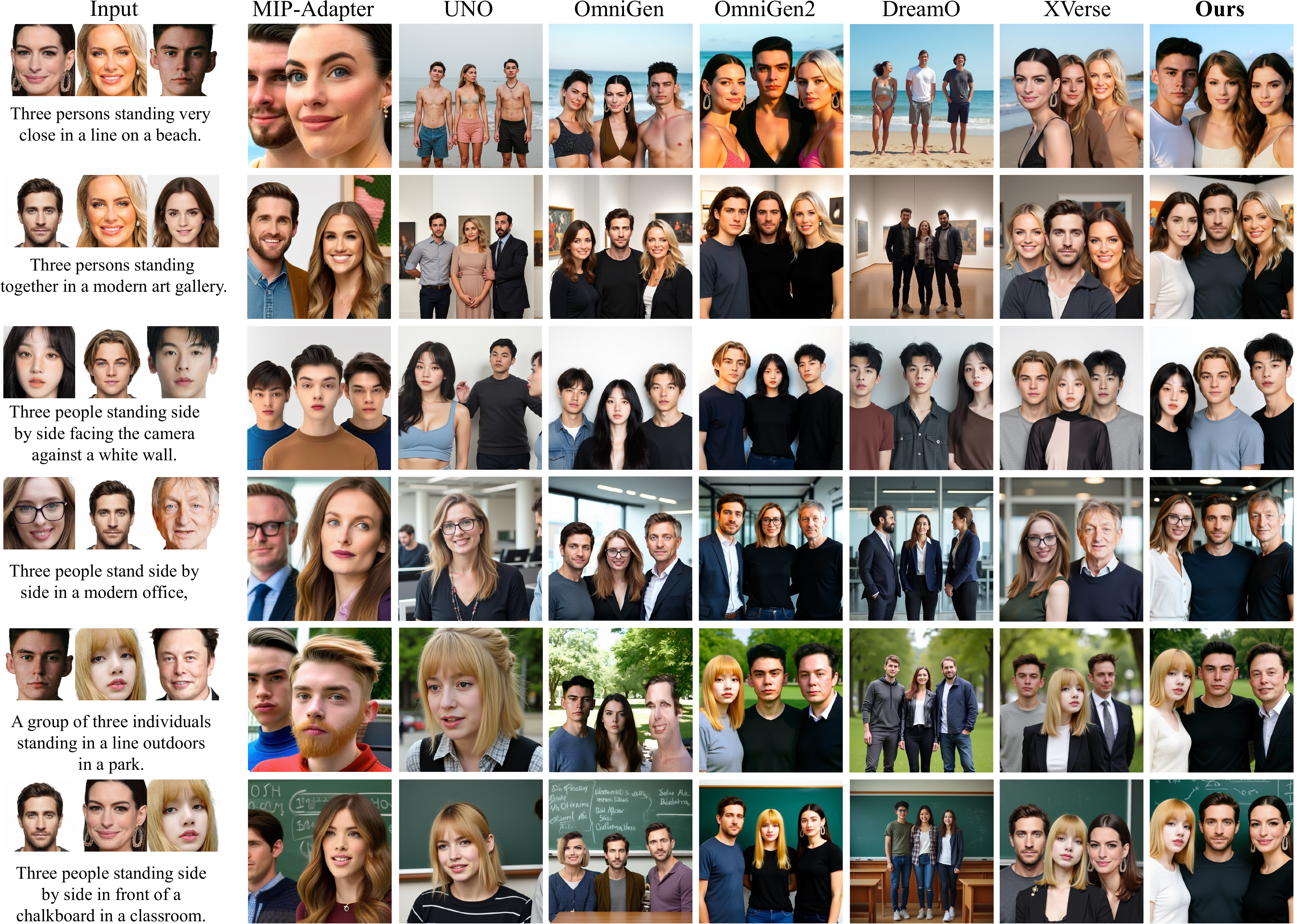}
    \caption{More Visualization of our method in three-human Generation.}
    \label{fig:cmp_three_human}
\end{figure*}

To further demonstrate our method's performance in multi-human personalization, we present qualitative comparisons in \Cref{fig:cmp_multi_human1} and \Cref{fig:cmp_multi_human2}. The results show that our model effectively preserves the identity of each subject and avoids the "attribute leakage" common in other methods. 
This outcome validates the efficacy of our Identity-Disentangled Attention Regularization (IDAR). 
While some baselines produce more stylized outputs that may yield higher HPSv2 scores, this is often at the expense of subject fidelity. 
Our method prioritizes photorealism and faithful subject appearance consistency, which leads to more reliable results in multi-subject customization.
Furthermore, we extended our method to scenarios involving more subjects to validate the effectiveness of our framework. We curated 10,000 samples of three-person customized generation data from video data to train with our method. As shown in \Cref{fig:cmp_three_human}, our method continues to significantly improve subject fidelity compared to existing methods, while maintaining text alignment and image aesthetics, effectively aligning with multi-dimensional human preferences. This demonstrates the effectiveness of our framework.


\section{More Results of Multi-Object Generation.}
\label{sec:app_multi_object}

\begin{figure*}[htb]
    \centering
    \includegraphics[width=1.0\linewidth]{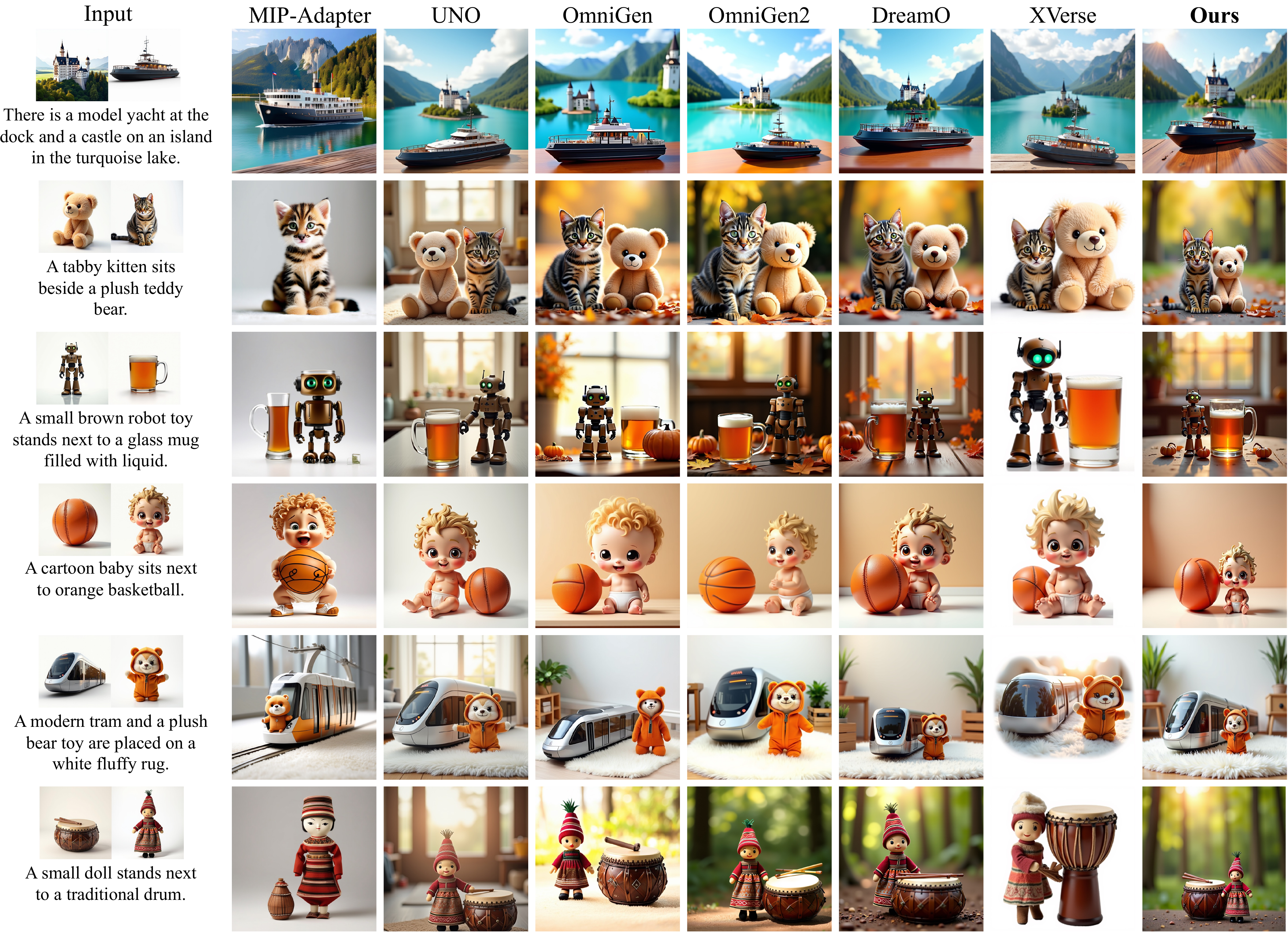}
    \caption{Visualization of our method in Multi-object Generation.}
    \label{fig:cmp_multi_object}
\end{figure*}

To evaluate the generalization of our framework, we showcase multi-object customization comparisons in \Cref{fig:cmp_multi_object}.
Our method demonstrates high object fidelity, accurately preserving core visual attributes such as a toy's texture or a glass's geometry. 
In contrast, competing approaches often introduce artifacts like deformation and detail loss. 
This highlights our model's strength in precise subject representation rather than hyper-stylization, a crucial capability for practical applications that require accuracy.

\section{More Results of Single-Subject Generation.}
\label{sec:app_single_human}

\begin{figure*}[htb]
    \centering
    \includegraphics[width=1.0\linewidth]{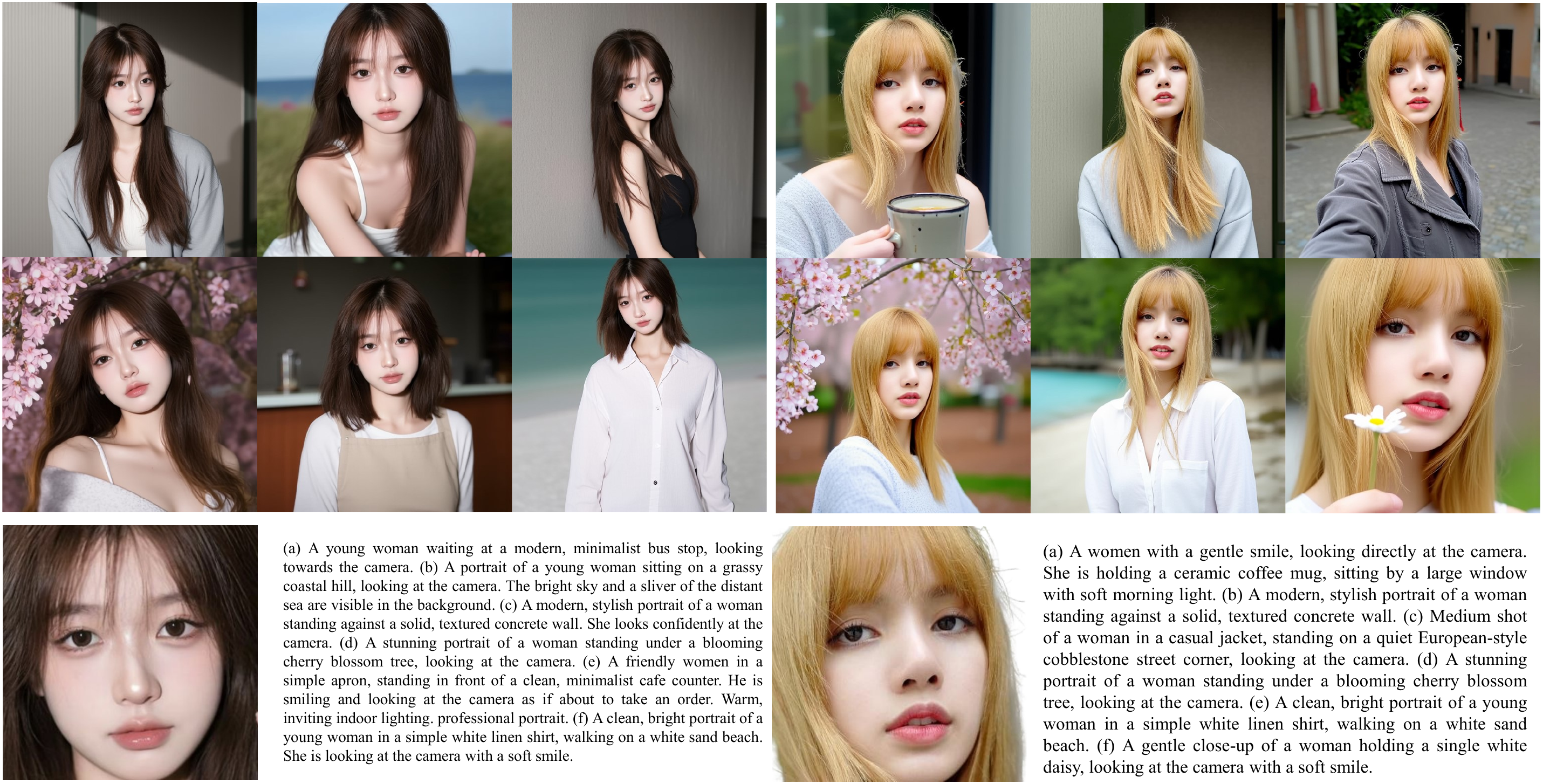}
    
    \vspace{2mm} 
    
    \includegraphics[width=1.0\linewidth]{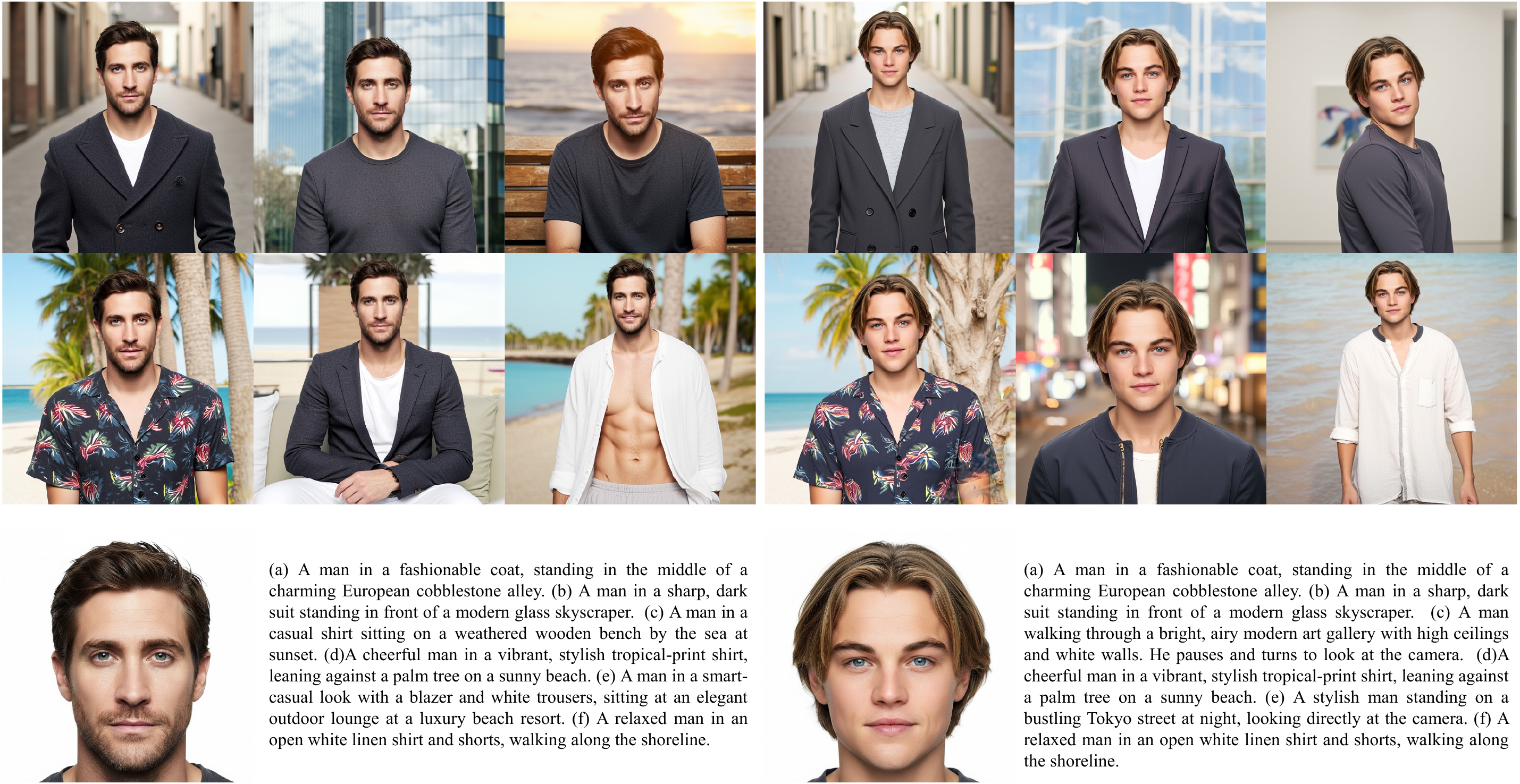}
    
    \caption{Visualization of our method in Single-subject generation. }
    \label{fig:single_merged_vertical}
\end{figure*}
\begin{figure*}[htb]
    \centering
    \includegraphics[width=1.0\linewidth]{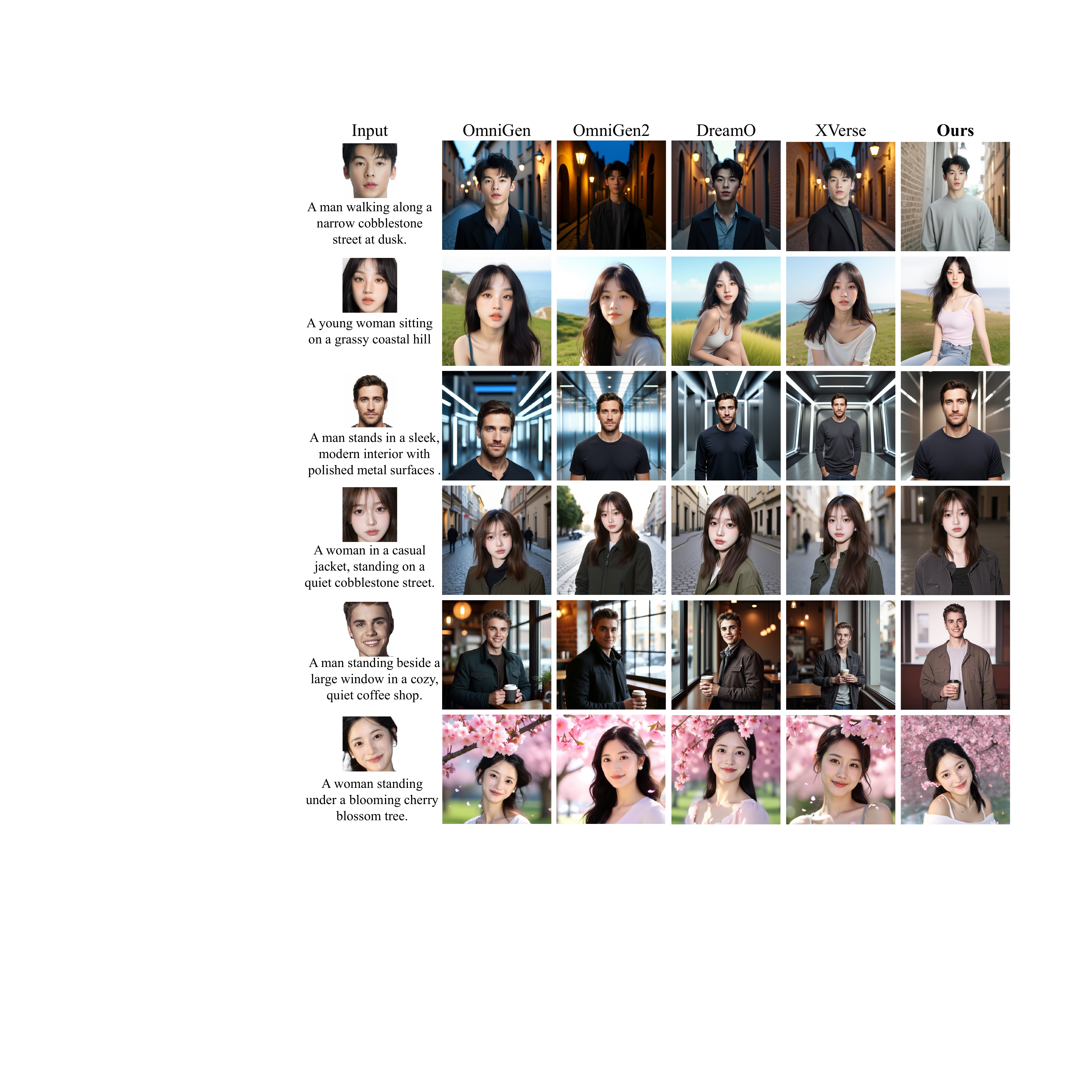}
    \caption{Qualitative comparison with existing methods on the single human generation.}
    \label{fig:cmp_single_human}
\end{figure*}

Effective multi-subject generation builds on strong single-subject performance. 
We validate this capability in \Cref{fig:single_merged_vertical} and \Cref{fig:cmp_single_human}, showing six diverse samples for each of four individuals and six frontal single-subject comparisons against SOTA models. 
Our method consistently preserves identity across varying styles, poses, and scenes, and even improves fidelity to the reference over baselines. 
These results confirm that the proposed framework not only enables reliable multi-subject generation but also enhances single-subject identity fidelity.


%% file: tables/data_analysis.tex
\begin{table}[t]
    \centering
    \caption{Analysis of Data Quality and Preference Alignment. We compare the original UNO with a version retrained on our dataset (UNO$^\dag$). Training with UNO on our data also caused the aesthetic metrics to drop to levels similar to those before we used IPPO.}

    \label{tab:data_ablation}
    \resizebox{1.0\linewidth}{!}{
    \begin{tabular}{l c ccc c}
        \toprule
        \textbf{Methods} & CLIP-T $\uparrow$ & Face-Sim $\uparrow$ & DINO-I $\uparrow$ & CLIP-I $\uparrow$ & AES $\uparrow$ \\
        \midrule
        UNO (Official) & 0.2645 & 0.1474 & 0.5972 & 0.6489 & 0.2954 \\
        UNO (Retrained)$^\dag$ & 0.2658 & 0.2879 & 0.7173 & 0.7528 & \cellcolor{badcolor}0.2656 \\
        \midrule
        Ours (w/o IPPO) & 0.2674 & 0.5154 & 0.8107 & 0.8480 & \cellcolor{badcolor}0.2661 \\
        \textbf{Ours (Full)} & \textbf{0.2753} & \textbf{0.5284} & \textbf{0.8294} & \textbf{0.8524} & \cellcolor{goodcolor}\textbf{0.2915} \\
        \bottomrule
    \end{tabular}
    }
    \vspace{-1em}
\end{table}